\documentclass[acmtog,nonacm]{acmart}
\acmSubmissionID{188}

\usepackage{booktabs} 

\usepackage{subcaption}
\usepackage{graphicx}
\usepackage{algpseudocode}
\usepackage{booktabs} 
\usepackage{etoolbox}
\usepackage{subcaption}
\usepackage{multirow}
\usepackage{tikz}
\usetikzlibrary{positioning}
\usepackage{tabularx}
\usepackage{dblfloatfix}
\usepackage{svg}

\newcommand{\mydraft}{false}
\usepackage[ruled]{algorithm2e} 

\renewcommand{\tt}{\mathbf{t}}
\newcommand{\BB}{\mathbf{B}}

\newcommand{\fm}{\boldsymbol{\phi}}

\newcommand{\ww}{\mathbf{w}}
\newcommand{\CC}{\mathbf{C}}

\newcommand{\xx}{\mathbf{x}}

\newcommand{\yy}{\mathbf{y}}

\newcommand{\XX}{\mathbf{X}}
\newcommand{\WW}{\mathbf{W}}
\newcommand{\RR}{\mathbf{R}}
\newcommand{\TT}{\mathbf{T}}

\newcommand{\js}{\boldsymbol{\theta}}

\SetAlFnt{\small}
\SetAlCapFnt{\small}
\SetAlCapNameFnt{\small}
\SetAlCapHSkip{0pt}

\algnewcommand\algorithmicforeach{\textbf{for }}
\algdef{S}[FOR]{For}[1]{\algorithmicforeach\ #1\ \algorithmicdo}

\acmJournal{TOG}
\acmArticle{39}

\begin{document}
\title{Shallow Signed Distance Functions for Kinematic Collision Bodies}

\begin{teaserfigure}
	\centering
	\includegraphics[draft=\mydraft,width=\textwidth,trim={5px 5px 5px 5px},clip]{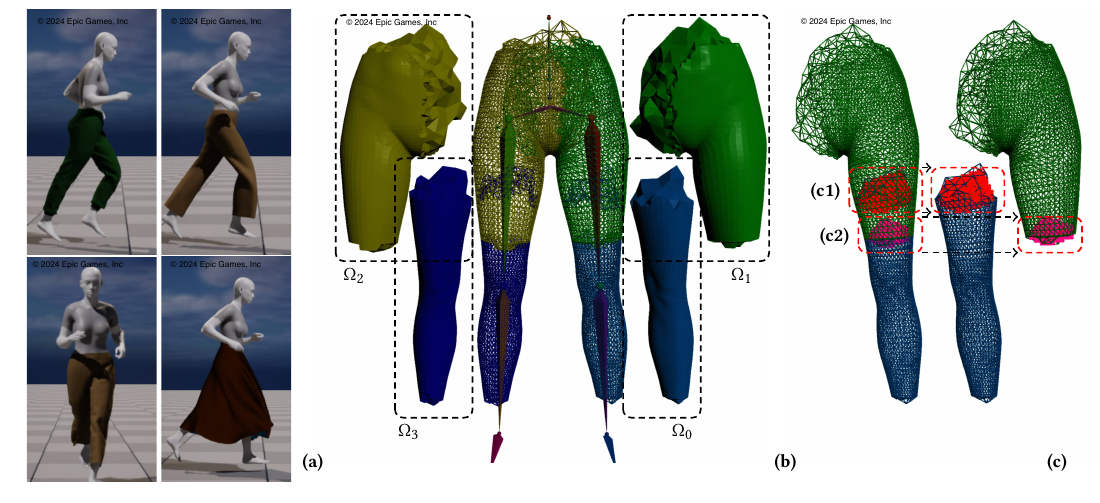}
	\caption{\textbf{Overview}. \textbf{(a)} Learning based SDFs are used in cloth simulations in real time.  \textbf{(b)} Our method partitions the character domain into subregions, and each region is represented by very fast shallow MLP neural networks. \textbf{(c)} Combining multiple SDFs requires additional information if the queried point is closer to interior boundary or the correct boundary. \textbf{(c1,c2)} highlights the region for the knee and the thigh in which the points inside are closer to the interior boundary than the avatars' boundary. The regions are determined by our neural network.}
	\label{fig:teaser}
\end{teaserfigure}

\author{Osman Akar}
\affiliation{\institution{UCLA}	\country{USA}}
\affiliation{\institution{Epic Games, Inc} \country{USA}}
\email{oak@math.ucla.edu}

\author{Yushan Han}
\affiliation{\institution{UCLA}\country{USA}}
\affiliation{\institution{Epic Games, Inc}\country{USA}}
\email{yushanh1@math.ucla.edu}

\author{Yizhou Chen}
\affiliation{\institution{UCLA}\country{USA}}
\affiliation{\institution{Epic Games, Inc}\country{USA}}
\email{chenyizhou@ucla.edu}

\author{Weixian Lan}
\affiliation{\institution{UC Davis}\country{USA}}
\email{wlan@ucdavis.edu}

\author{Benn Gallagher}
\affiliation{\institution{Epic Games, Inc}\country{USA}}
\email{benn.gallagher@epicgames.com}

\author{Ronald Fedkiw}
\affiliation{\institution{Stanford University}\country{USA}}
\affiliation{\institution{Epic Games, Inc}\country{USA}}
\email{fedkiw@cs.stanford.edu}

\author{Joseph Teran}
\affiliation{\institution{UC Davis}\country{USA}}
\affiliation{\institution{Epic Games, Inc}\country{USA}}
\email{jteran@math.ucdavis.edu}

\begin{abstract}
We present learning-based implicit shape representations designed for real-time avatar collision queries arising in the simulation of clothing.
Signed distance functions (SDFs) have been used for such queries for many years due to their computational efficiency.
Recently deep neural networks have been used for implicit shape representations (DeepSDFs) due to their ability to represent multiple shapes with modest memory requirements compared to traditional representations over dense grids.
However, the computational expense of DeepSDFs prevents their use in real-time clothing simulation applications. 
We design a learning-based representation of SDFs for human avatars whoes bodies change shape kinematically due to joint-based skinning.
Rather than using a single DeepSDF for the entire avatar, we use a collection of extremely computationally efficient (shallow) neural networks that represent localized deformations arising from changes in body shape induced by the variation of a single joint.
This requires a stitching process to combine each shallow SDF in the collection together into one SDF representing the signed closest distance to the boundary of the entire body.
To achieve this we augment each shallow SDF with an additional output that resolves whether or not the individual shallow SDF value is referring to a closest point on the boundary of the body, or to a point on the interior of the body (but on the boundary of the individual shallow SDF).
Our model is extremely fast and accurate and we demonstrate its applicability with real-time simulation of garments driven by animated characters. 
\end{abstract}

%
%


%
%


\maketitle

\section{Introduction and Related Work}
Simulation of deformable objects is ubiquitous in modern computer graphics applications.
Whether it is the intricate stretching and folding of cloth, or the squash, stretch and contraction of soft tissues in virtual characters, elastic deformation is essential for creating satisfactory visual realism in modern visual effects and video games.
Simulation of this type is complex and computationally expensive in general.
The most challenging aspects are generally self-collision detection/resolution \cite{bridson:2002:cloth,baraff:1998:cloth,wang:2014:dcc,wu:2020:cloth} and the rapid solution of large systems of nonlinear equations \cite{tang:2016:cama,wang:2018:multigrid,wang:2016:des_gpu}.
However, another important aspect is detection and resolution of collision constraints with kinematic geometric objects in the scene.
These kinematic objects are not influenced by the deformable objects in the scene (e.g. due to large mass ratios), however their motion often determines the deformation of the elastic objects of interest.
Clothing draping and interacting with a kinematic body shape is perhaps the most important example of this, and it is the focus of our approach.\\
\\
In most simulation techniques, elastic object collision with kinematic collision bodies is imposed as a constraint on the governing equations.
These constraints are detected and enforced using a variety of geometric descriptions of the collision body.
We focus on the use of machine learning techniques for accelerating this process.
Although many recent methods have investigated the use of these techniques in clothing simulation, most replace the simulation process altogether with a neural network.
While fast, learning techniques are still limited in accuracy compared to simulation, particularly in the case of free flowing cloth with significant inertia.
Our aim is to avoid these limitations by replacing just the kinematic body collision portion of a typical cloth simulation pipeline with machine learning enhancement.
Our proposed approach is unique in this way, however we briefly discuss a number or recent techniques that utilize machine learning in relevant ways.
Romero et al. \shortcite{romero:2023:mlc} use a neural network to learn a displacement mapping for resolving elastic object collisions against kinematic rigid bodies with reduced models for deformable objects.
Tan et al. \shortcite{tan:2022:repulsive} note that purely-learning based cloth techniques suffer greatly from self and kinematic body collision artifacts. 
Santesteban et al. \shortcite{santesteban:2021:self} also address this problem by adding a repulsion term into their loss function so that trained cloth will be less likely to penetrate the body.
Betiche et al. \shortcite{bertiche:2020:cloth3d,bertiche:2021:deepsd} also add body-collision based loss terms into training to discourage cloth/body penetration.
Gundogdu et al. \shortcite{gundogdu:2019:garnet} do this as well.\\
\\
Signed distance functions (SDFs) generally have constant query time (e.g. when pre-computed and stored over dense grids) and are very useful when rapidly detecting and resolving collisions between a kinematic animated body and simulated clothing \cite{osher:2003:LSMDIS}.
However, SDF calculation (e.g. over dense grid nodes) is expensive and is therefore usually done as an offline/pre-computation.
Furthermore, while pre-computed SDFs over regular grids are effective, there are some notable drawbacks. 
First, SDFs are usually pre-computed at frame-rate time intervals since sub-frame time steps would require excessive computation of the SDFs (even as a pre-computation).
Temporal interpolation of SDFs can be used for sub-frame time queries \cite{selle:2008:msm}.
Dynamic time step sizes (e.g. resulting from CFL conditions with explicit or semi-implicit time stepping) cannot be known a priori and require this sub-frame interpolation of precomputed SDFs (or excessive non-pre-computation of SDFs on the fly).
Also, the storage cost of SDFs at each frame in an animation for each character in the scene quickly becomes excessive.
Lastly, the motion of the character must be known completely before the simulation is carried out if pre-computation is to be used.
While this is a reasonable assumption for some applications (e.g. offline visual effects), it is not possible in real-time simulation environments where the user is actively redefining the character motion on the fly.
\\
\\
Recently, a variety of neural network models for SDFs have been proposed.
\cite{Ortiz:etal:iSDF2022, Koptev_Neural_ISDF} use neural networks to approximate signed distance functions for scene reconstruction for robotics in real time.\cite{genova2019learning} learn a general shape template from data.
\cite{metasdf_neurips_20} use meta-learning to perform the same task as \cite{park:2019:cvpr}, representing multiple 3D shapes.
\cite{Liu_2023_CVPR} use unsigned distance function for shape construction for volume rendering.
Ma et al. \shortcite{NeuralPull} use neural networks to learn SDF representations of point clouds.
Chabra et al. \shortcite{chabra:2020:deep} utilize local shape patches to increase the variety of shapes representable with neural SDFs.
The Deep SDF approach of Park et al. \shortcite{park:2019:cvpr} is particularly powerful.
In this case, a network is trained to represent a discrete collection of shapes by training over their individual SDFs (sampled over regular grids).
Each shape is encoded with a representative shape vector in the process.
These functions can represent a wide range of shapes and utilize dramatically less memory than a collection of SDFs defined over regular grids.
In the context of representing the body shape of a kinematic animated character, DeepSDFs could be used to model the SDF of the skin surface rigged with joint-based skinning (e.g.  linear blend skinning) over a {\textit{discrete}} collection of joint states.
However, real-time simulation in this context requires collision queries against the shape of the kinematic avatar skin at {\textit{continuous}} samples of the joint state since animation states cannot be discretely sample a priori.\\
\\
To enable application of learning-based SDF techniques in real-time clothing simulation, we design a neural network SDF that depends continuously on both the collision query point and the kinematic joint-state vector.
Rather than using a single DeepSDF defined over the entire body, we use a collection of extremely shallow and computationally efficient networks that represent the skin surface very accurately near individual joints.
This joint-local approach efficiently focuses network degrees of freedom where they are needed and allows for additive scaling complexity of training data burden (in the number of joints).
That is, each joint network can be trained separately without the need to couple the effect of distant joints.
However, by decoupling into joint-centric shallow SDFs we lose some information about the signed distance to the surface of the complete skinned character since each joint SDF refers to only a portion of the character.
This means that the joint-centric SDF zero-isocontours may coincide with the true boundary or may coincide with an internal boundary specific to the joint.
We correct for this by training our networks to know whether or not the signed-distance value is associated with a true boundary or an internal boundary.
This knowledge allows us to blend the join-centric SDFs into an efficient and accurate SDF for the skin of the character.
\\
\\
Linear blend skinning (LBS) \cite{thalmann:1989:lbs} is an effective and widely-used means for defining the skin surface of animated characters from a kinematic joint state.
However, the LBS surface is not guaranteed to be self-intersection free which complicates the definition of a SDF representation (e.g. near joints with large ranges of motion like the elbow and knee).
We compensate for this by training on surfaces that have had LBS self-collisions resolved in a simulation post-process.
We demonstrate the accuracy and efficiency of our approach with real-time simulation of clothing colliding against representative animated skin surfaces of human avatars.
In summary, our contributions can be listed as:
\begin{itemize}
\item Learning-based SDFs that vary continuously with the kinematic joint state of animated characters.
\item Shallow joint-centric neural networks trained to represent local skin deformation.
\item A boolean variable returned by each joint-centric shallow SDF that indicates whether a query point is associated with a fictitious joint-internal surface or the global skin boundary.
\item A blending mechanism for computing the SDF to the union of the regions defined by each joints shallow SDF.
\item Resolution of self-collision artifacts in the SDF of LBS surfaces.
\end{itemize} 

\section{Character Kinematics}\label{sec:char}
We assume that the kinematics of the animated character are define by a joint-state vector $\js\in\mathbb{R}^{\hat{N}_J}$ where $\hat{N}_J$ denotes the number of joint degrees of freedom in the animation rig.
We use $\js_i\in\mathbb{R}^{D_i}$, $1\leq i \leq {N}_{J}$ to denote the individual the joint-state vector components where $1 \leq D_i \leq 3$ depending on the type of joint.
For simple pin joints, $D_i=1$ but for a general revolute joint $D_i=3$.
Note that we did not consider the fully-general case of a 6-degree of freedom rigid joint in this work.
With this convention we have
$\js=\left(
\js_1,
\hdots,
\js_{{N}_{J}}
\right)^T\in\mathbb{R}^{\hat{N}_J}$
where $\hat{N}_J=\sum_{i=1}^{{N}_J}D_i \leq 3N_J$.
We further assume that the kinematics of the character motion are defined in terms of a deformation mapping $\fm:\Omega\times\mathbb{R}^{\hat{N}_J}\rightarrow\mathbb{R}^3$ where $\Omega\subset\mathbb{R}^3$ is the three dimensional domain of the interior of the character in a reference pose.
We use $\Omega^{\js}=\left\{\xx\in\mathbb{R}^3 | \exists \XX\in\Omega \ \textrm{such that} \ \fm(\XX,\js)=\xx \right\}$ to denote the interior region of the animated state of the body (given joint stata $\js$).
In our examples we define $\fm(\XX,\js)=\fm^\textrm{C}(\XX,\fm^\textrm{LBS}(\XX,\js))$ where $\fm^\textrm{LBS}(\XX,\js):\Omega\times\mathbb{R}^{\hat{N}_J}\rightarrow\mathbb{R}^3$ is the standard linear blend skinning and $\fm^\textrm{C}:\Omega\times\mathbb{R}^{3}\rightarrow\mathbb{R}^3$ is a collision corrective that resolves collision/pinching in $\fm^\textrm{LBS}$ so that $\fm(\cdot,\js):\Omega\rightarrow\Omega^{\js}$ is always bijective.
The $\fm^\textrm{C}$ corrective on LBS is defined in Section~\ref{sec:mpm} and assures that the SDF of the animated state has ample room for clothing to surround the character body.
\\
\\
We partition the reference character domain $\Omega\subset\mathbb{R}^3$ into subregions $\Omega_i\subset\Omega$ associated with each joint $0\leq i <N_J$ such that $\Omega=\cup_{i=1}^{N_J} \Omega_i$. 
Each subregion $\Omega_i$ is the portion of the interior of the character (in the reference pose) that is most deformed by changes in the joint state.
Generally, each $\Omega_i$ is determined based on proximity to the joint transformation center.
See Section~\ref{sec:mpm} for further discussion.
Note that these subsets are not intersection free in general $\Omega_i\cap\Omega_j \neq 0$ as nearby joints will influence similar regions.
We use $\Omega_i^{\js}=\left\{\xx\in\mathbb{R}^3 | \exists \XX\in\Omega_i \ \textrm{such that} \ \fm(\XX,\js)=\xx \right\}$ to denote the animated state of the joint sub-region.

\section{Signed Distance Function}\label{sec:sdf}
\begin{figure}[h]
	\centering
	\includegraphics[draft=\mydraft, width=0.32\linewidth, trim={18px 0 20px 0},clip]{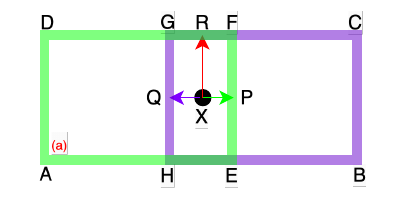}
	\includegraphics[draft=\mydraft, width=0.32\linewidth, trim={18px 1px 20px 0},clip]{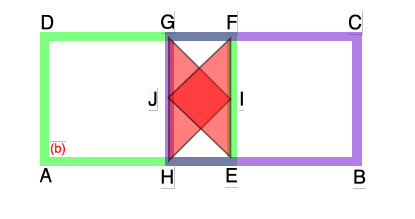}
	\includegraphics[draft=\mydraft, width=0.32\linewidth, trim={18px 0 20px 0},clip]{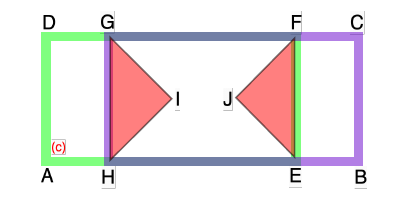}
	\caption{Example of combinining two SSDFs into one. The rectange $ABCD$ is divided into two parts $AEFD$ and $BHGC$. \textbf{(a)} Point $X$ is closer to the interior boundary than the true boundary for both SSDFs. \textbf{(b)} Red regions $JFE$ and $HIG$ shows incorrect boundary information for $AEFD$ and $BHGC$. The two partitions above lead to undesired regions where the correct SDF cannot be computed. \textbf{(c)} Our partition is analogous to AEFD and BHGC here so that the incorrect boundary regions (red) do not intersect, therefore SDF can be computed for all interior points.}
	\label{fig:dz_explained}
\end{figure}

We define the signed distance to the surface of the animated character as $\phi:\mathbb{R}^3\times\mathbb{R}^{\hat{N}_J}\rightarrow\mathbb{R}$.
Here $|\phi(\xx,\js)|$ denotes the distance from a point $\xx\in\mathbb{R}^3$ to the closest point on the skin surface of the character in the animated state defined by the joint state vector $\js$. 
The sign of $\phi(\xx,\js)$ indicates whether the point $\xx$ is inside the skin surface or outside.
The closest point on the skin surface to the point $\xx$ is determined as $\xx-\phi(\xx,\js)\nabla^\xx \phi(\xx,\js)$ where $\nabla^\xx \phi(\xx,\js)$ is the gradient of $\phi(\xx,\js)$ with respect to $\xx$.
We define the signed distance function ${\phi}(\xx,\js)$ in terms of a collection of joint-wise augmented signed distance functions $\boldsymbol{\phi}_i:\mathbb{R}^3\times\mathbb{R}^{\hat{N}_J}\rightarrow\mathbb{R}\times\mathbb{B}$.
For each joint $i$, these return the signed distance $\phi_i(\xx,\js)$ from the point $\xx$ to the boundary of the animated region $\Omega_i^{\js}$ associated with the joint as well as a boolean $b_i(\xx,\js)\in\mathbb{B}$ that indicates whether the closest point to $\xx$ on the boundary of $\Omega_i^{\js}$ is on the true boundary of animated character $\Omega^{\js}$ ($b_i(\xx,\js)=1$) or whether it is on the interior of $\Omega^{\js}$ ($b_i(\xx,\js)=0$).
A portion of the boundary of $\Omega_i^{\js}$ coincides with the boundary of $\Omega^{\js}$ and another portion is interior to $\Omega^{\js}$ and this boolean is used to resolve the true signed distance when a point $\xx$ is in multiple joint subregions.
Note that $b_i(\xx,\theta_i)=1$ for all $\xx$ that are not on the interior of $\Omega^{\js}$.
See Figure~\ref{fig:dz_explained} for an illustration of the boolean values and choice of the subregions to avoid undesired cases.
We use the notation $\boldsymbol{\phi}_i(\XX,\theta_i)=(\phi_i(\XX,\theta_i),b_i(\XX,\theta_i))$ to represent our augmented signed distance convention.
With this formalism, the signed distance function is defined as
\begin{align}
\phi(\xx,\js)=\min_{i \in S(\xx,\js)} \phi_i(\xx,\js)\label{eq:b_phi}
\end{align}
where $S(\xx,\js)\subset\left\{0,1\hdots,N_J-1\right\}$ is the collection of joint indices $i$ such that $b_i(\xx,\js)=1$. 
Note that $S(\xx,\js)\ne \emptyset$, for any $\xx$ there is at least one subregion so that closest point from $\xx$ to $\Omega_i$ is on the true boundary. We enforce this by defining the subregions to be analogous to Figure~\ref{fig:dz_explained}(c).

\section{Shallow Joint Signed Distance Functions}
\begin{figure}[h]
	\centering
	\includegraphics[draft=\mydraft, width=\linewidth, trim={0px 730 0px 0},clip]{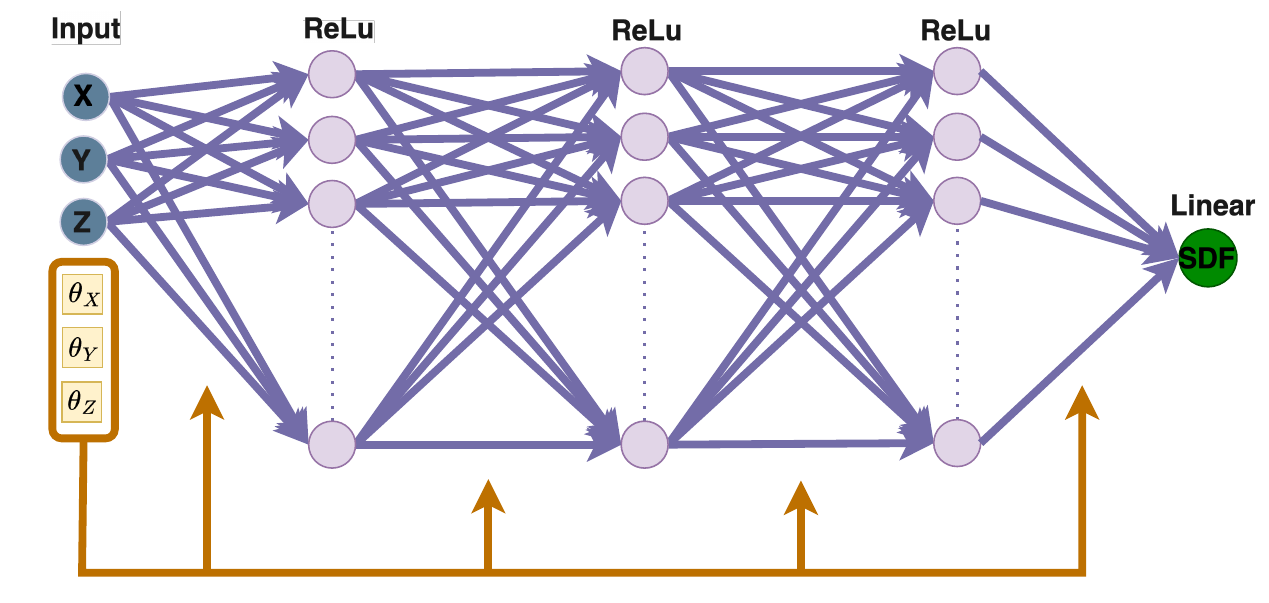}
	\caption{\textbf{Model Architecture}. Weights $\{\WW^{n}_{ij}\}$ and biases $\{b^{n}_{ij}\}$  between the $n^{th}$ and $(n+1)^{th}$ layers for channel $l$ are computed as in Equation~\ref{eq:weights}. The figure illustrates the architecture for $N_J=5$ (3 hidden layers) and each hidden layer has $n=8$ activation nodes. The hidden layers have rectified linear unit (ReLU) activation function, the output layer has linear activation function. This illustration assumes the joint has $D_i=3$ degrees of freedom: The rotation angles in $X,Y,Z$ direction. }
	\label{fig:ML_model_diagram}
\end{figure}

\begin{figure}[h]
	\centering
	\includegraphics[draft=\mydraft, width=0.49\linewidth, trim={1450px 500 1400px 800}, clip]{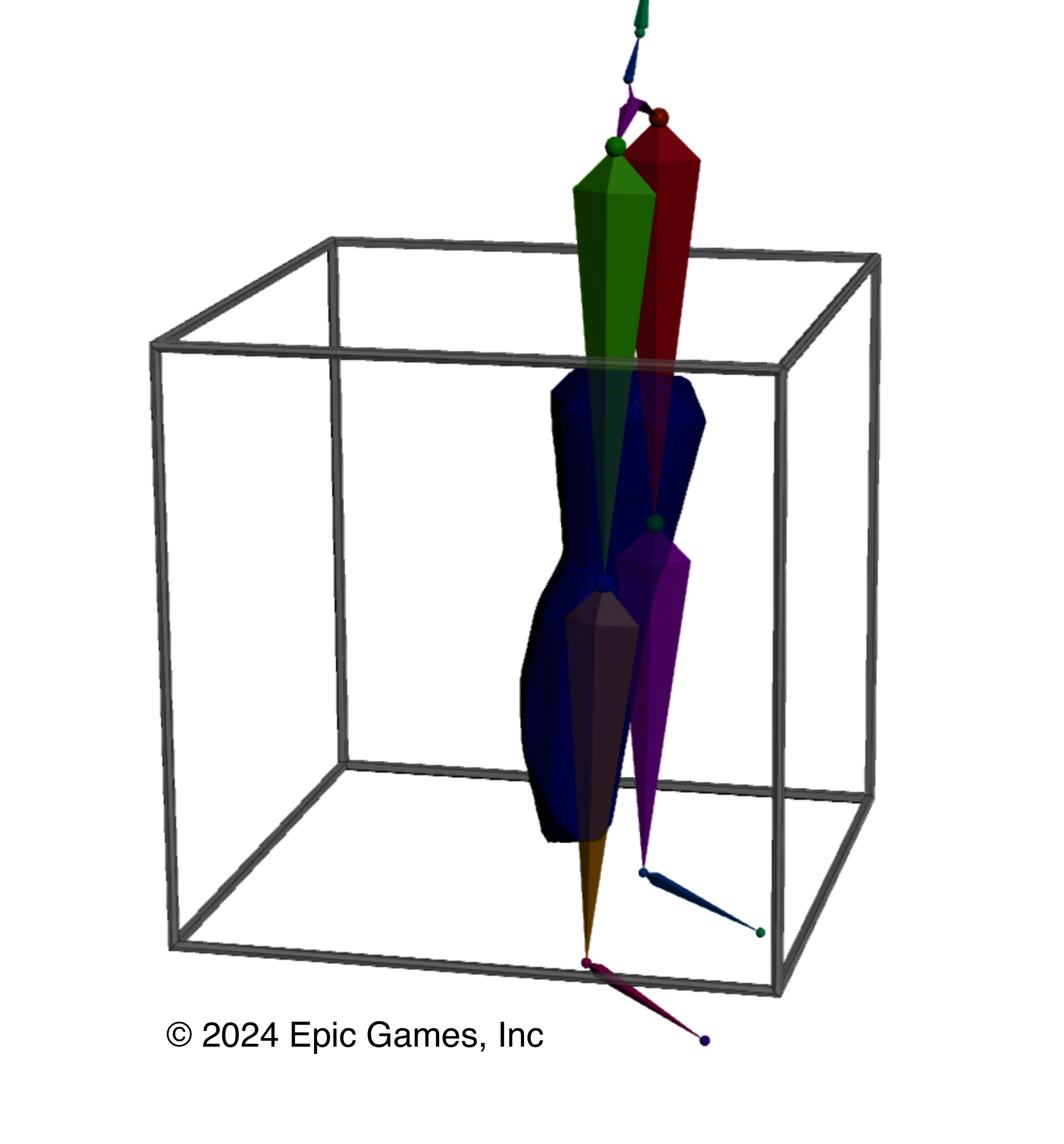}
	\includegraphics[draft=\mydraft, width=0.49\linewidth, trim={1650px 500 1200px 800}, clip]{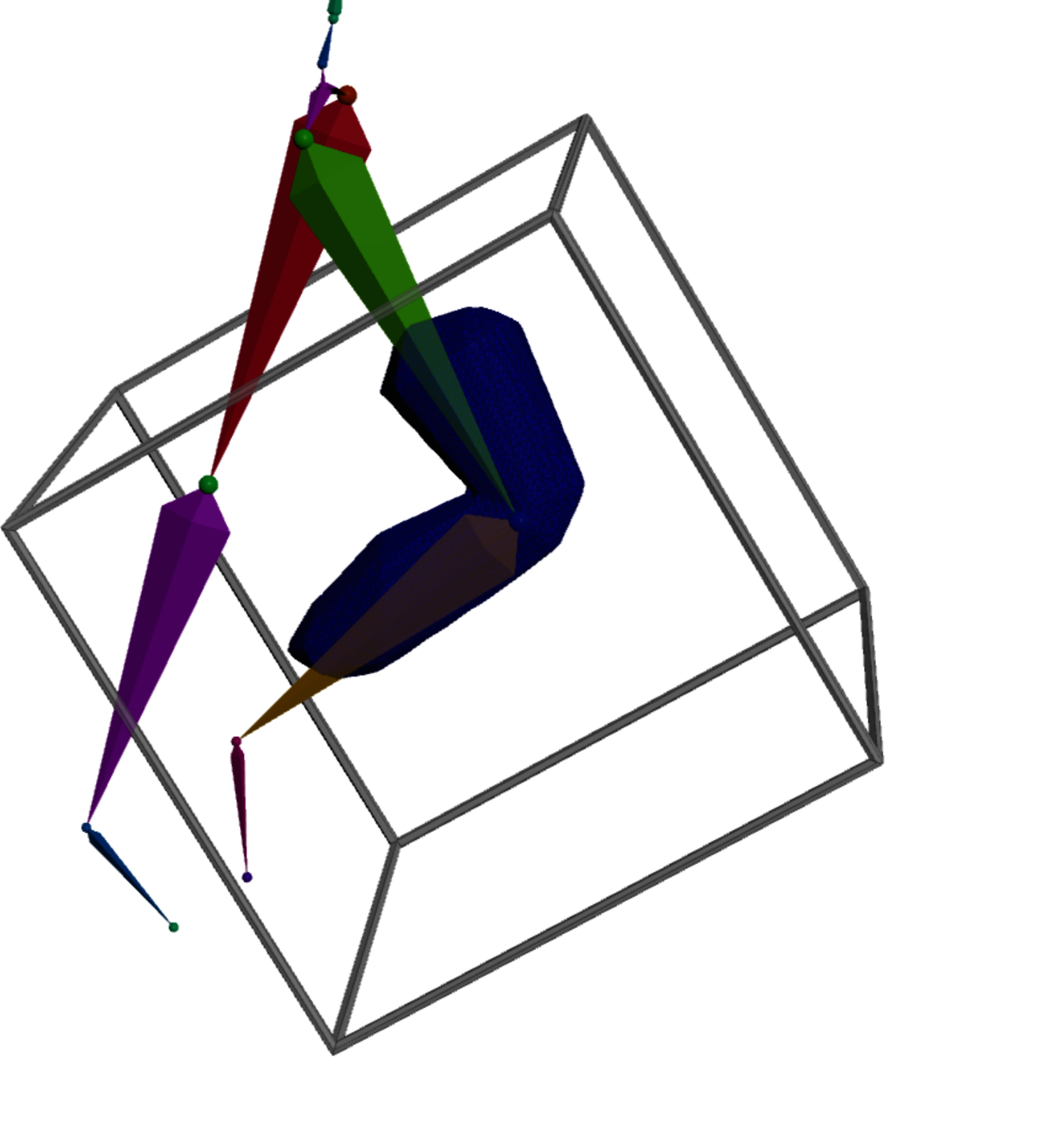}
	\caption{The canonical space of SSDF for the joint knee is determined by thigh, the parent of knee in the skeletal hierarchy.}
	\label{fig:parent_grid}
\end{figure}

We define each joint-wise signed distance function $\boldsymbol{\phi}_i:\mathbb{R}^3\times\mathbb{R}^{N_J}\rightarrow\mathbb{R}\times\mathbb{B}$ in terms of shallow neural networks that can be evaluated efficiently in real-time and are accurate enough to represent the deformed shape of the animated joint region $\Omega_i^{\js}$.
To enhance the ability of our neural network parameters to capture the signed distance values over a range of animated states, we find it helpful to define each signed distance in the canonical space associated with each joint.
We define this space in terms of joint-wise rigid transforms $\TT_i(\xx,\js)=\RR_i(\js)\xx+\tt_i(\js)$ for rotations $\RR_i(\js)$ and translations $\tt_i(\js)$ as well as a Multilayer Perceptron (MLP, \cite{haykin:1994:neural}) neural network $SSDF:\mathbb{R}^3\times\mathbb{R}^{N_W}\times\mathbb{R}^{N_B}\times\mathbb{R}^{D_i}\rightarrow\mathbb{R}$ where $N_W$ is the number of weights and $N_B$ is the number of biases as
\begin{align}
\phi_i(\xx,\js)=SSDF(\TT_i(\xx,\js),\WW_i,\CC_i,\js_i).\label{eq:s_phi}
\end{align}
The transform $\TT_i(\xx,\js)$ is chosen according to the parent transform joint in the hierarchy so that deformation near the joint is isolated from rigid motion in the rig.
See Figure~\ref{fig:parent_grid} for illustration.
We find that allowing the effective weights and biases in each layer to depend linearly on the joint degrees of freedom $\js_i=\left(\theta_{i1},\hdots,\theta_{iD_i}\right)^T$ as
\begin{align}\label{eq:weights}
\hat{\ww}^n_{ij}(\js_i)=\sum_{\alpha=1}^{D_i} \ww^n_{ij\alpha}\theta_{i\alpha} + {\ww}^n_{ij0}, \ \hat{c}^n_{ij}(\js_i)=\sum_{\alpha=1}^{D_i} c^n_{ij\alpha}\theta_{i\alpha} + c^n_{ij0}
\end{align}
improved expressivity across ranges of joint rotations.
Here $n$ indicates that the weight connect layers $n$ and $n+1$ in the MLP (where $1 \leq n \leq N_L-1$) and $j$ refers to the channel in the layer $n+1$.
$N_L$ refers to the total number of layers in the MLP with the convention that all but the first (input) and the last (output) are the so called hidden layers.
$N_H$ refers to the number of channels in the hidden layers.
These feed into that activation functions $\sigma:\mathbb{R}\rightarrow\mathbb{R}$ to define 
$SSDF(\XX,\WW_i,\CC_i,\js_i)$ in terms of the per-layer channel values $y^{n+1}_j$ as
\begin{align}
	y^{n+1}_j=\sigma({\hat{\ww}_{ij}^n(\js_i)}^T\yy^n + \hat{c}^n_{ij}(\js_i)),\ 0 \leq n < N_L
\end{align}
where $\yy^1=(y^1_1,y^1_2,y^1_3)^T=\XX\in\mathbb{R}^3$ and $SSDF(\XX,\WW_i,\BB_i,\js_i)=y^{N_L}_1$.
See Figure~\ref{fig:ML_model_diagram} an illustration of this network structure. 
Note that with this convention, the weights relating the input and second (first hidden) layers have $\hat{\ww}^1_{ij}(\js_i)\in\mathbb{R}^3$, $1\leq j \leq N_H$.
For all other choices of $1<n<N_L$, $\hat{\ww}^n_{ij}(\js_i)\in\mathbb{R}^{N_H}$.
Furthermore for $1\leq n < N_L$, the channel index has $1\leq j \leq N_H$ however in the weights and biases connecting the last hidden layer with the output layer ($n=N_L-1$), there is only one output channel and the channel index is simply $j=1$.\\
\\
We use $\WW_i=\{\ww^n_{ij\alpha}\}\in\mathbb{R}^{N_W}$ and $\CC_i=\{b^n_{ij\alpha}\}\in\mathbb{R}^{N_B}$ Equation~\eqref{eq:s_phi} to represent the collection of all learnable weights and biases in the MLP network and note that the effective weights and biases in Equation~\eqref{eq:weights} are chosen in terms of them in this way to suit cloth simulation and collision against dynamic avatars.
In particular, the $SSDF$ is evaluated at each particle in the simulation mesh at each time step, but the dependence of the model on the joint state $\js$ happens only once over the time step.
The formula in Equation~\eqref{eq:weights} updates the weights whenever the joint state changes and once complete inference only with recomputes based on the positional inputs (with the effective weights held fixed).
This allows the network to have $(D_i+1)$ times more parameters in training, compared to what it is required at the inference.
In practice we used 3 hidden layers ($N_L=5$) and 8 channels per hidden layer ($N_H=8$).
\\

We use a similar network structure for the joint-wise boolean function
\begin{align}
b_i(\xx,\js)=bool(SSDF(\TT_i(\xx,\js),\tilde{\WW}_i,\tilde{\CC}_i,\js_i))\label{eq:bool}
\end{align}
and also evaluate it through the joint-local transform $\XX=\TT_i(\xx,\js)$.
Note that this network has its own learnable weights ($\tilde{\WW}_i$) and biases ($\tilde{\CC}_i$), however we used 2 hidden layers ($N_L=4$) and 8 channels per hidden layer ($N_H=8$).
Also note that $bool(\cdot)$ in Equation~\eqref{eq:bool} returns false for negative values of the neural network and true for positive values.

\section{Training and Dataset Creation}\label{sec:dataset}
\begin{figure}[h]
	\begin{tikzpicture}
		\node [anchor=south west, inner sep=0pt] (image) at (0,0) {\includegraphics[draft=\mydraft, width=1\linewidth, trim={1300px 700 1300px 500},clip]{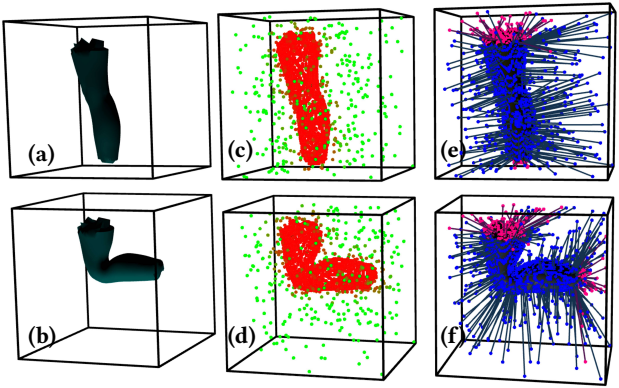}};
		\begin{scope}[x={(image.south east)},y={(image.north west)}]
			\node [anchor=north west, inner sep=1pt, text=black] at (0.05,0.64) {\textbf{(a)}};
			\node [anchor=north west, inner sep=1pt, text=black] at (0.05,0.18) {\textbf{(b)}};
			\node [anchor=north west, inner sep=1pt, text=black] at (0.36,0.65) {\textbf{(c)}};
			\node [anchor=north west, inner sep=1pt, text=black] at (0.36,0.18) {\textbf{(d)}};
			\node [anchor=north west, inner sep=1pt, text=black] at (0.69,0.65) {\textbf{(e)}};
			\node [anchor=north west, inner sep=1pt, text=black] at (0.69,0.18) {\textbf{(f)}};
		\end{scope}
	\end{tikzpicture}	
	\caption{Left: Deformation of knee surrounded by the training grid. Upper image is the rest state (0,0), below image is (0,-90). Note that for knee there are only two degrees of freedom. Middle: Points on the grid selected for training data. The color of the node represents the distance to the boundary (red: close, green: distant). Right: Blue points have correct signed distance as the closest point lies on the origional boundary, the purple points have incorrect signed distance values.}	
	\label{fig:dataset_with_DZ}
\end{figure}
To train the joint-wise neural networks we first partition the reference pose of the body $\Omega$ (which we take to be an A-pose with the characters arms at their side) into the joint-based regions $\Omega_i$.
We assume that the character skin surface vertices have weights associated with each transform in the rig so that LBS can be applied.
We then tetrahedralize the interior of the skin surface and generate skinning weights on newly created interior tetrahedron vertices by solving a Poisson equation.
Dirichlet boundary conditions are applied on the surface of the tetrahedron mesh and set to the values of the LBS weights.
We then associate all vertices with weight values above a threshold with the parent transform of each joint to define the seed region for the $\Omega_i$.
We then assign vertices that did not have a weight above a threshold for any $\Omega_i$ greedily to regions associated with vertices connected in the tetrahedron mesh.
This region growing is continued until all vertices are assigned to a region $\Omega_i$.
We then grow each region slightly to make sure that there is sufficient overlap to apply the logic of Figure~\ref{fig:dz_explained}(c).
Figure~\ref{fig:teaser} illustrates this process in a representative example.\\
\\
To train the SSDF for each $\Omega_i$, grid based SDFs are generated for a range of joint poses. 
For a particular joint pose $\js_i$, $\Omega_i$ is deformed to $\Omega_i^{\hat{\js}_i}$ (where $\hat{\js}_i\in\mathbb{R}^{N_J}$ has all joint variables except $\js_i$ set to the A-pose) where the mesh-based elastic material point method (MPM) of Jiang et al. \shortcite{jiang:2015:apic} is used to prevent any collisions associated with LBS.
Specifically, the vertices of the tetrahedron mesh overlapping bones in the rig are tracked with Dirichlet displacement boundary conditions in a quasistatic elasticity solve using mesh-based MPM.
This defines the LBS collision correction mapping in $\fm^C$ in Section~\ref{sec:char}.
Once this collision-free version of the LBS mapping is defined, we define SDF values over a regular grid using exact geometric distances on cut grid cells which are swept to the remaining grid nodes using the Fast Marching Method \cite{sethian:1996:fast}.\\
\\
For the joint $i$ with $D_i$ degrees of freedom, we create training range as the product space of evenly spaced values.
More precisely, the training range for joint $i$ is the product space of  $\bigotimes_{\alpha=1}^{D_i}[\js_{i\alpha;min}:\js_{i\alpha;inc}:\js_{i\alpha;max}]$
where $[\js_{i\alpha;min}:\js_{i\alpha;inc}:\js_{i\alpha;max}]$ is $$\{\js_{i\alpha;min}+k*\js_{i\alpha;inc}|0\le k \le \frac{\js_{i\alpha;max}-\js_{i\alpha;min}}{\js_{i\alpha;inc}}\}$$
Here all angles are in euler angles.
For the examles we provide, the following range of motion are used:\\ 
Knees: 2 degrees of freedom with product space $[-20:10:20]\bigotimes[-150:10:30]$. Note that we are not using the rotation input that associates with twisting, which does not make much difference for the clothing other than very tight clothes.\\
Thighs: 2 degrees of freedom with product space $[-80:10:10]\bigotimes[-90:10:90]$.\\
\\
We use $100\times100\times100$ uniform grid and compute signed distance values for all $100^3$ nodes for each pose and use them to generate the training data.
The grid can be thought as a bounding box that covers subregion $\Omega_i$ for all deformations of $\js_i$ in the training range with a $10\%$ buffer.
Note that the points near the 0-isocontour are the most important points to determing the 0-levelset.
We devise a probabilistic selection method to create the training data from the grid based SDFs (we do not use all the points).
Point $\xx$ is selected to appear in the dataset for joint $i$ if either $|SDF[\js_i](\xx)|$<$\epsilon$ or $|SDF[\js_i](\xx)|* \mathbf{I} < \beta$.
Here $\mathbf{I}$ is a uniform random variable distributed over interval [0,1], $\epsilon$ is the boundary selection bound, and $\beta$ is the  randomized selection bound. \\
\\
For our training data, we choose $\epsilon=0.025*L_G$ and $\beta = 0.001*L_G$ where $L_G$ is the side length of the bounding box.
Figure~\ref{fig:dataset_with_DZ}-(b,e) shows an example of chosen grid points.
This probabilistic process selects about 30000 grid nodes among 100000.
As it can be seen from the Figure~\ref{fig:dataset_with_DZ}, all points near boundary are selected, whereas the points further away from boundary have less chance to be chosen.
Furthermore, another ml model is trained to determine if a selected point is closer to the interior boundary or not. 
For each selected point, we assign a float value 1 if the closest point is on the correct boundary and -1 if on the interior boundary. Figure~\ref{fig:dataset_with_DZ}(c,f) shows labeling for knee.\\
We use a modified version of clamped loss function as suggested in \cite{park:2019:cvpr}. $$L(\phi_i(\xx,\js_i),s)=|clamp((\phi_i(\xx,\js_i),\delta)) - clamp(s,\delta)|^2$$
where $s$ is the ground truth signed distance value and clamping function is defined as $clamp(s,\delta)=min\{\delta,max\{-\delta,s\}\}$.
Smaller clamping values allows network to focus on the boundary.
In our experiments we saw that $L2$ error creates visually better results compared to $L1$.
We choose $\delta=0.2*L_G$.
We train our model with TensorFlow \cite{abadi:2015:tensorflow} on a single NVIDIA RTX A6000 GPU with 48GB memory.
We use $3GB$ of a shared memory, allowing us to train multiple models at once.
Training is done with back-propagation and the ADAM \cite{Kingma2015AdamAM} optimizer with learning rate 0.001.
We train our network for 100K epochs. 
Figure~\ref{fig:lower_body_multiple_epochs} shows the 0-levelset of the SSDFs after 1K, 10K, 50K and 100K epochs.
Training takes approximately 4 hours for 100K epochs for our network.

\begin{figure}[h]
	\centering
	\includegraphics[draft=\mydraft, width=0.9\linewidth, trim={0 0 0 0 }, clip]{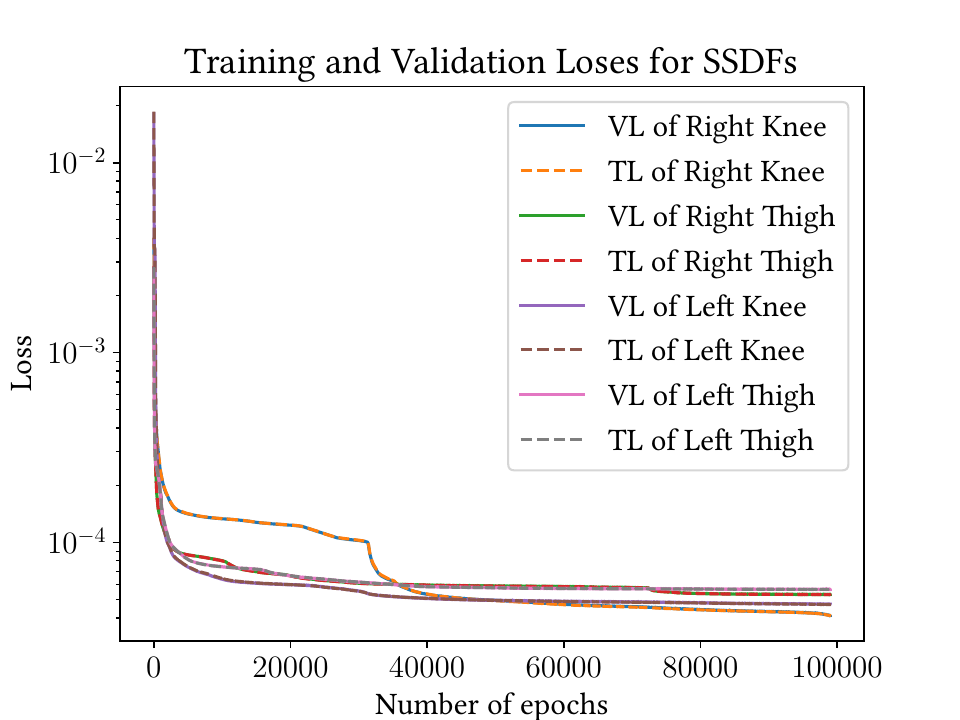}
	\caption{Training and validation losses for the Shallow SDF networks for subregions. $y-axis$ is log scaled.}
	\label{fig:training_losses}
\end{figure}

\begin{figure}[h]
	\centering
	\includegraphics[draft=\mydraft, width=0.9\linewidth, trim={200px 100 160px 200}, clip]{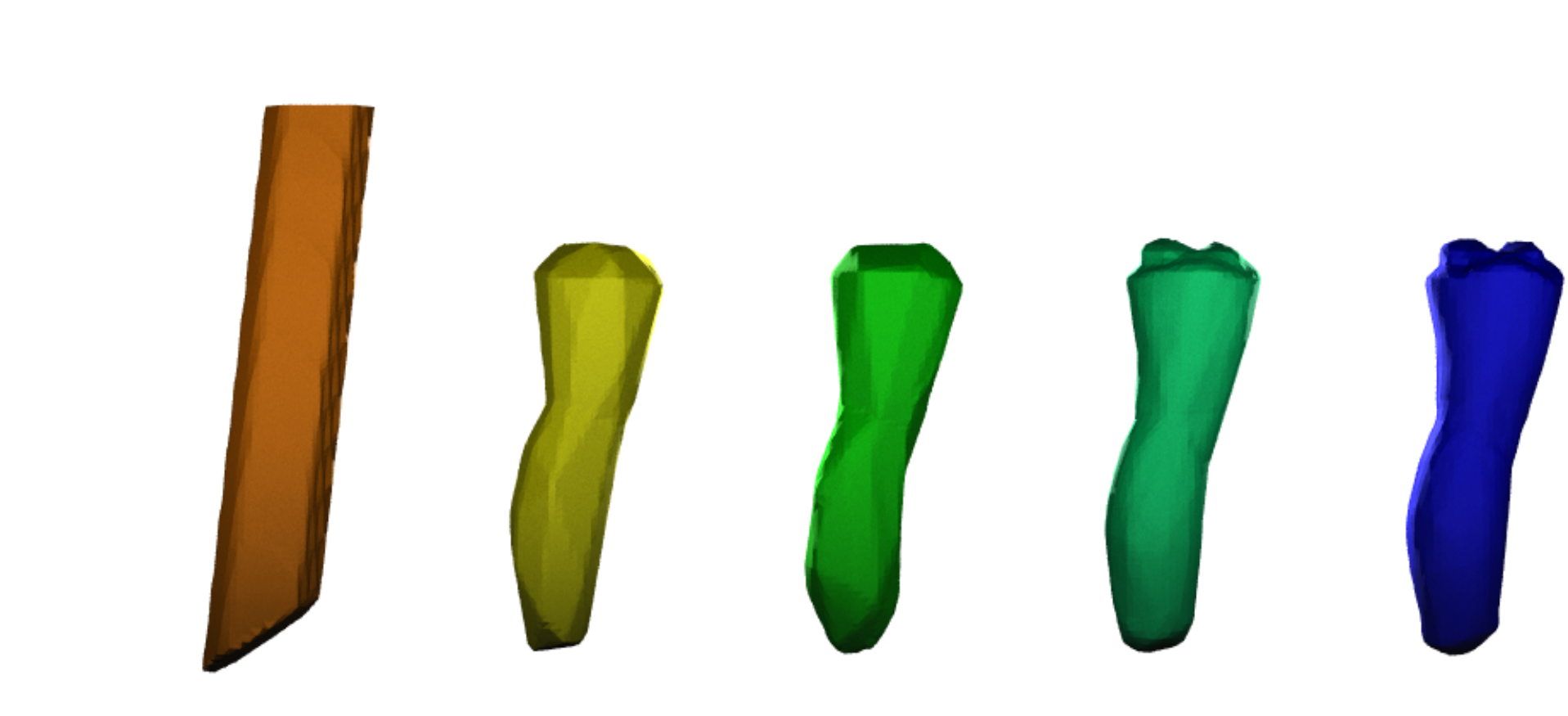}
	\includegraphics[draft=\mydraft, width=0.9\linewidth, trim={200px 200 160px 200}, clip]{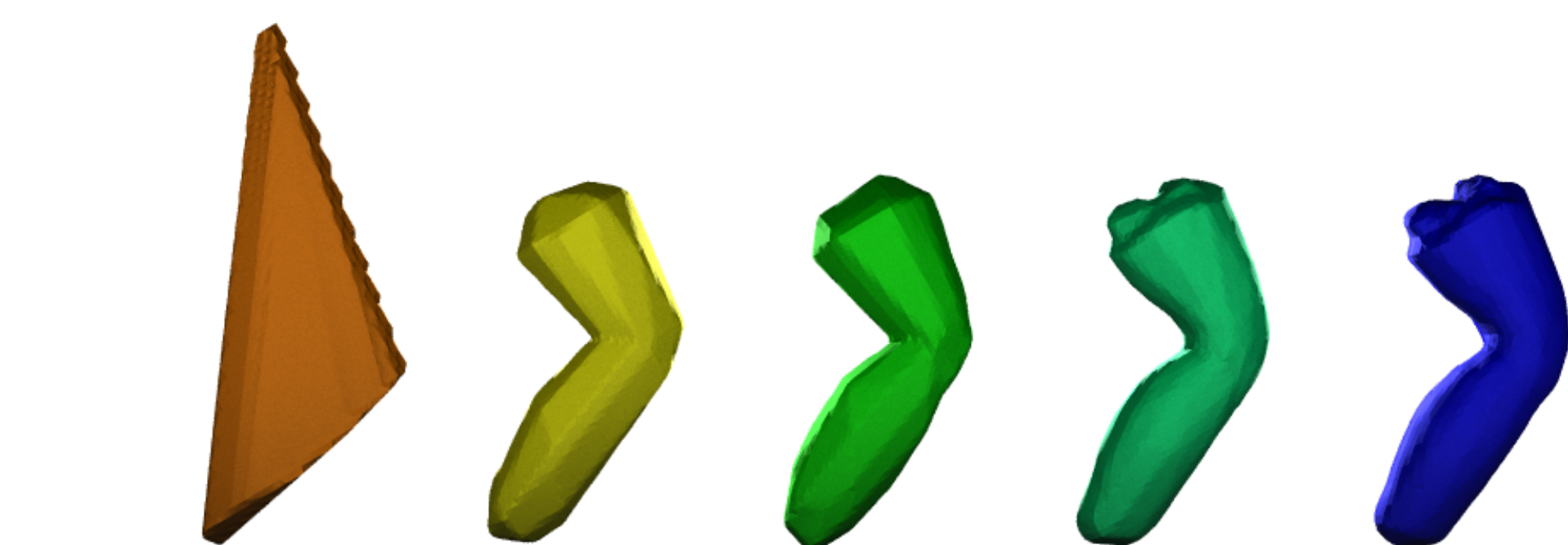}
	\includegraphics[draft=\mydraft, width=0.9\linewidth, trim={200px 200 160px 300}, clip]{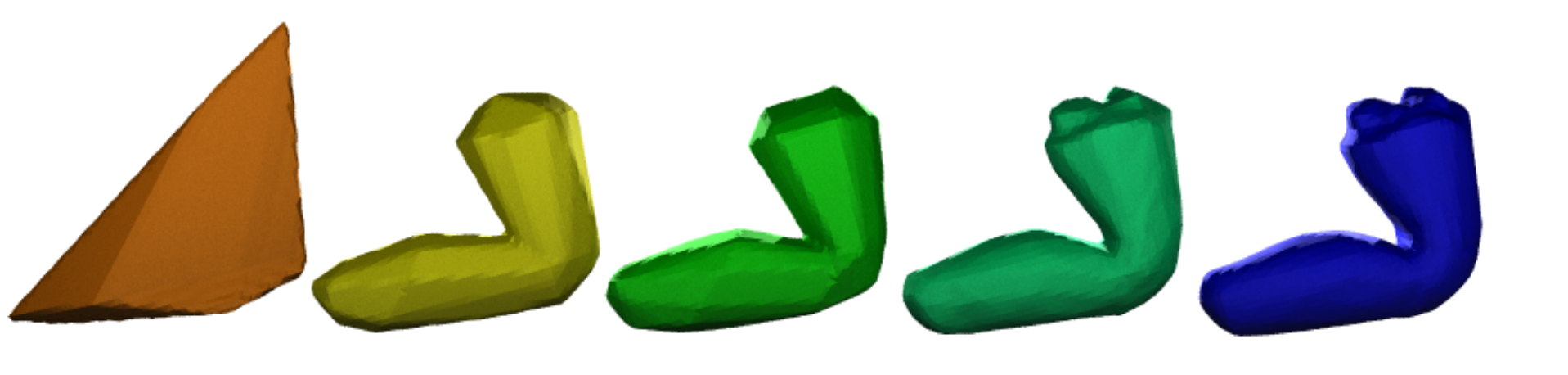}
	\caption{Example of SSDFs with different network structures trained for the right knee. Top to bottom 0-levelset of the deformed object for three different joints states are illustrated. From left to right the following network structures are used: $(N_L=5, N_H=4)$, $(N_L=5, N_H=8)$, $(N_L=5, N_H=16)$, $(N_L=7, N_H=2)$ and $(N_L=7, N_H=32)$. All networks are trained for 100K epochs. We choose the network structure in yellow for the balance between speed and performance.}
	\label{fig:calf_single_diff_models}
\end{figure}

\begin{table}[t]
	\caption{\textbf{Training and Evaluation Loss}.}
	\resizebox{\columnwidth}{!}{%
		\begin{tabular}{@{}llllll@{}}
			\toprule
			Network Choices & $P_T$ & $P_I$ & $L_T$ & $L_V$&$T_{train}$  \\ \midrule
			$N_L=5, N_H=4$ & $183$ & 61 & $2.227\times10^{-4}$ & $2.222\times10^{-4}$ & ~3 hours\\
			$N_L=5, N_H=8$ & $555$ & 185 &$4.117\times 10^{-5}$  & $4.119\times10^{-5}$ &~4 hours\\
			$N_L=5, N_H=16$ & $1875$ &  625&$2.793\times 10^{-5}$ &  $2.812\times10^{-5}$ &~5 hours\\
			$N_L=5, N_H=32$ & $ 6819$ & 2273 &$5.518\times 10^{-6}$ &  $5.751\times10^{-6}$ &~7 hours\\
			$N_L=5, N_H=32$ & $13155$ & 4385 & $3.517\times 10^{-6}$ &  $3.505\times10^{-6}$ &~10 hours\\
			\bottomrule
		\end{tabular}
	}
	\label{tbl:network_structure}
	\caption{$P_T$ = Number of Parameters of the network for training. $P_I $ = Number of paramaters of the network for inference. $L_T$ = Training loss after 100K epochs. $L_V$ = Validation loss after 100K epochs. $T_{train}$ = Time it takes to train for 100K epochs. The results shown are for the SSDF of the right knee decribed in Figure~\ref{fig:calf_single_diff_models}. The degrees of freedom for the knee is $D_i=2$.} 
\end{table}

\section{Results and Examples}
\begin{figure}[h]
	\centering
	\includegraphics[draft=\mydraft, width=0.9\linewidth, trim={100px 250 200px 100},clip]{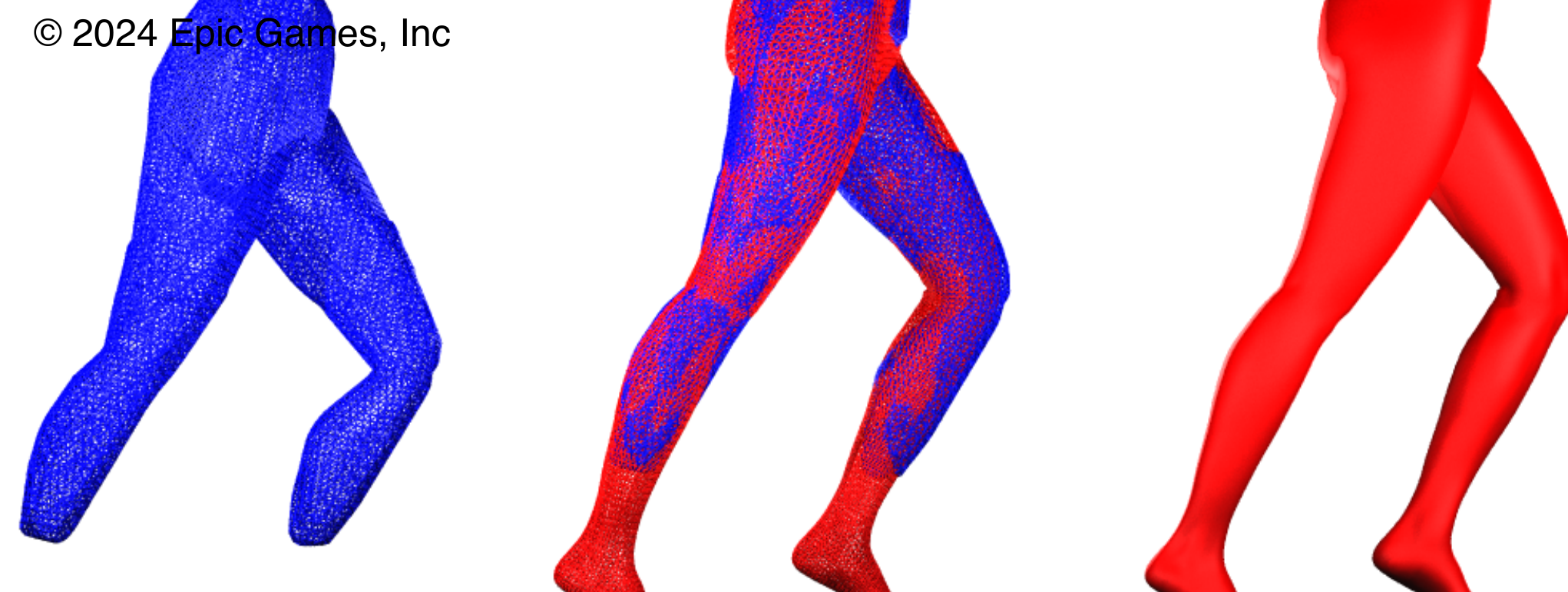}
	\includegraphics[draft=\mydraft, width=0.9\linewidth, trim={350px 100px 350px 118},clip]{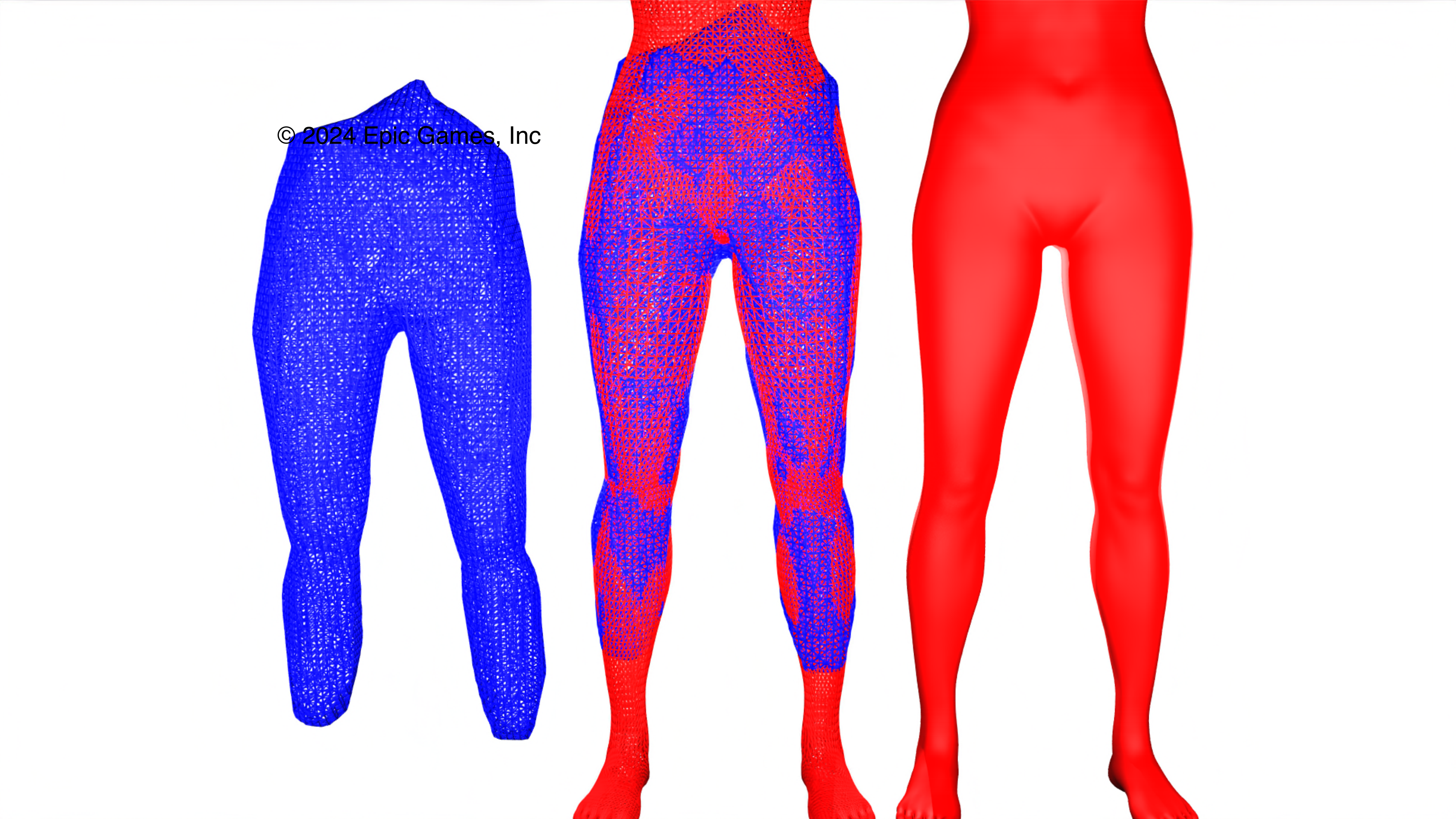}
	\includegraphics[draft=\mydraft, width=0.9\linewidth, trim={100px 250 200px 100},clip]{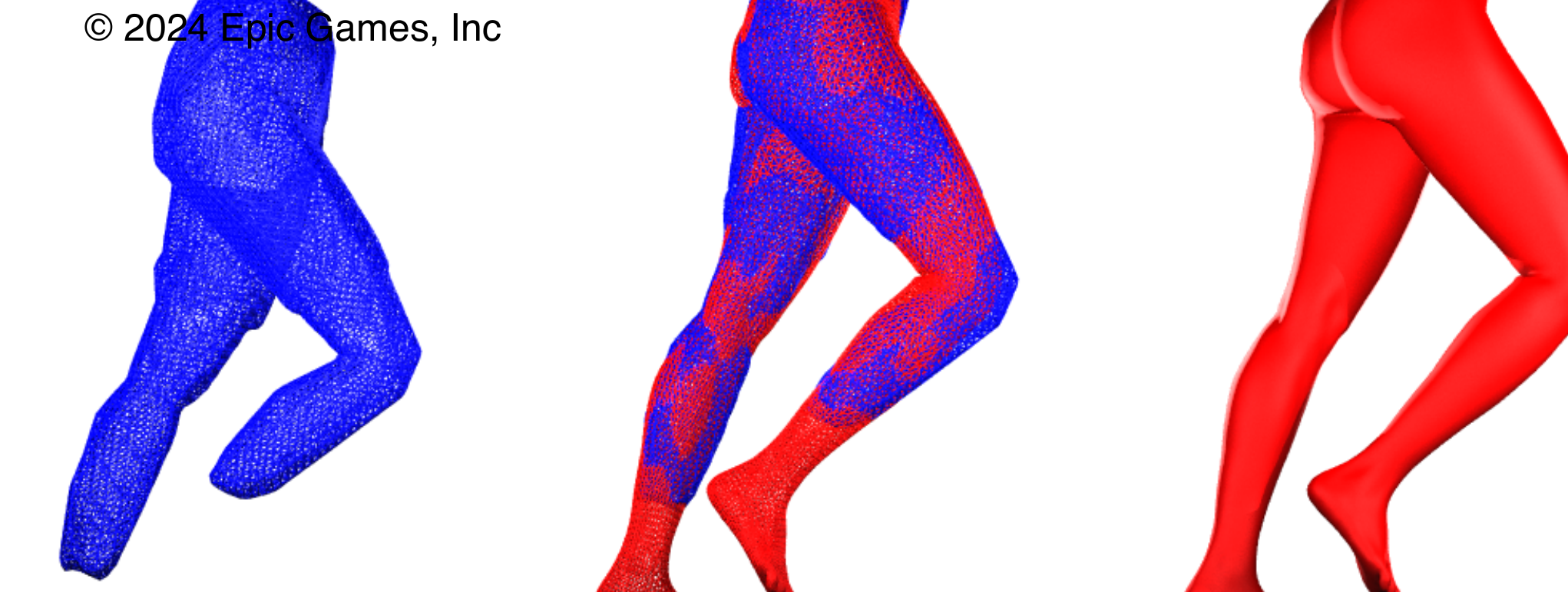}
	\caption{Zero-levelset derived from learned SDF with 4-SSDFs in 3 different joint states. From left to right: learned SDF, learned SDF and Ground Truth combined, and Ground Truth are presented.}
	\label{fig:iso_vs_gt_W8_d2}
\end{figure}

\begin{figure}[h]
	\centering
	\includegraphics[draft=\mydraft, width=0.95\linewidth, trim={100px 200 100px 150},clip]{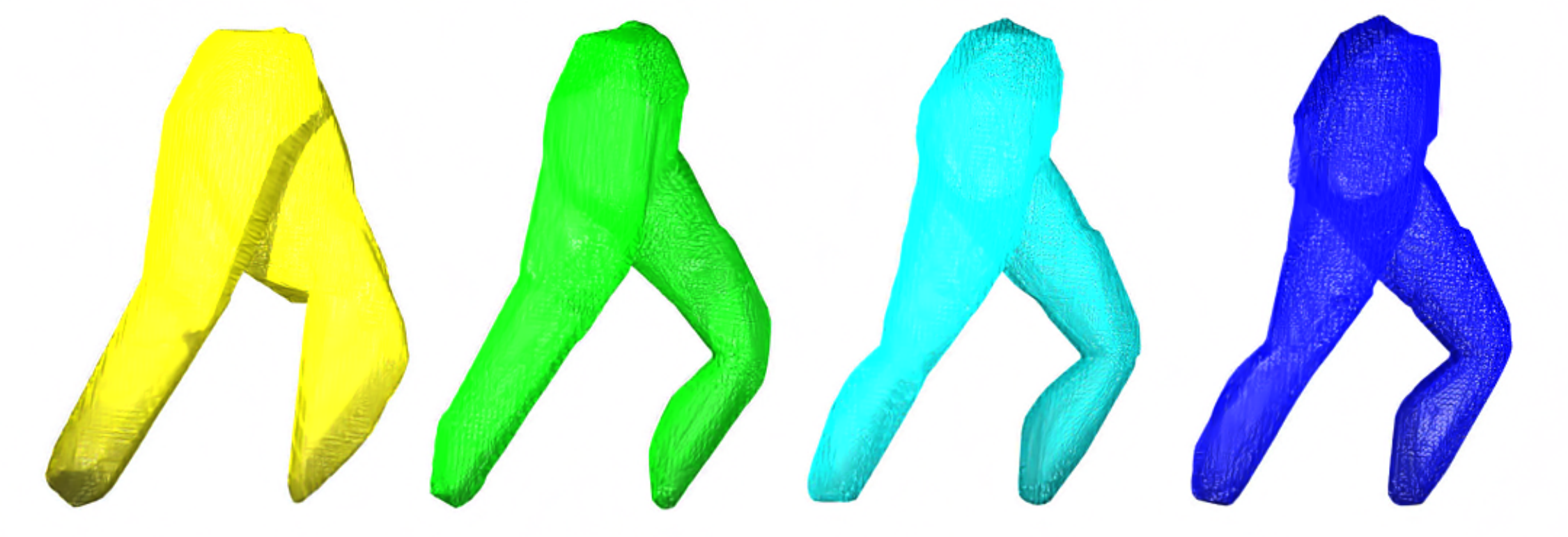}
	\includegraphics[draft=\mydraft, width=0.95\linewidth, trim={100px 200 100px 150},clip]{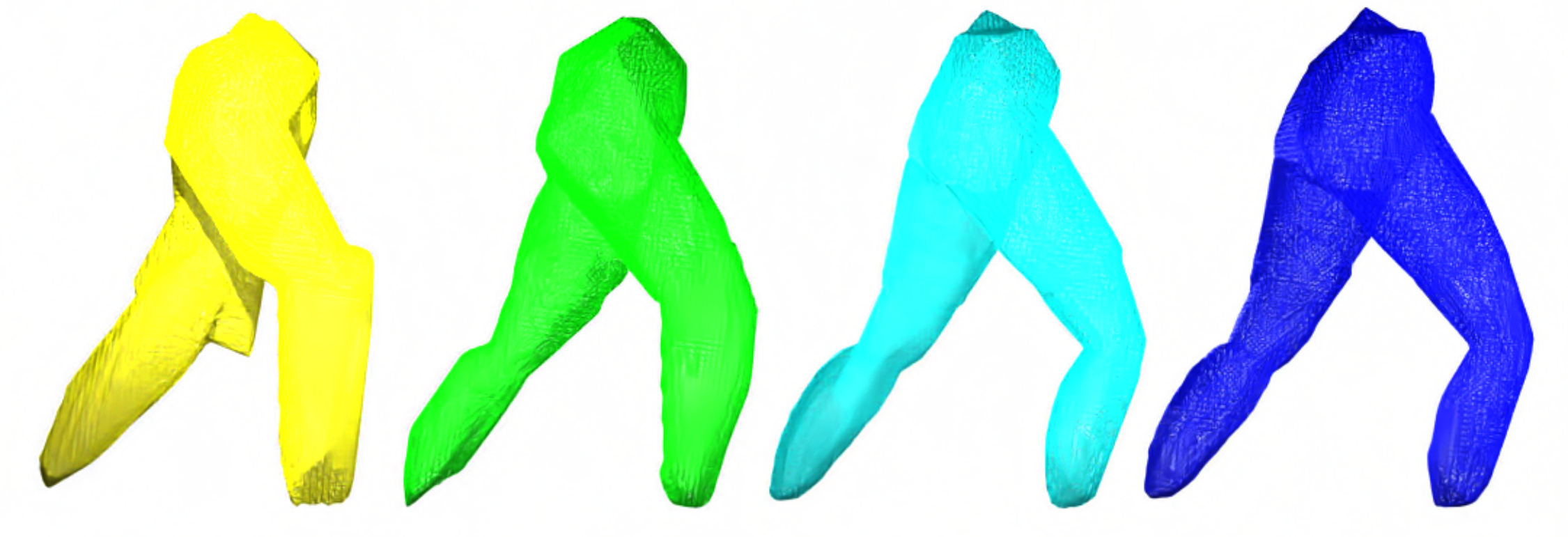}
	\caption{Zero-levelsets of the trained SDFs after 1K, 10K, 50K and 100K epochs.}
	\label{fig:lower_body_multiple_epochs}
\end{figure}

\begin{figure}[h]
	\centering
	\includegraphics[draft=\mydraft, width=0.24\linewidth, trim={600px 100 800px 0}, clip]{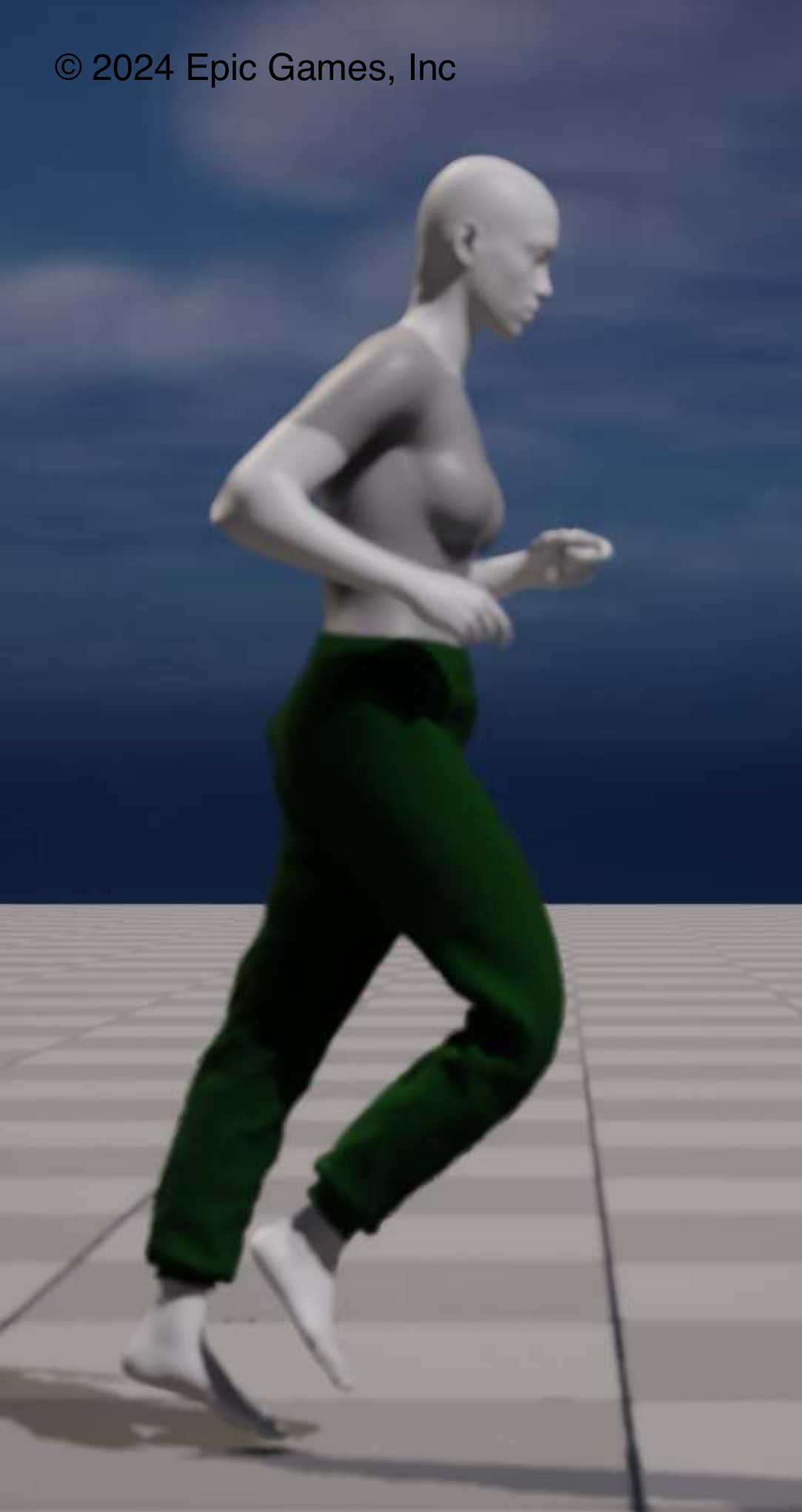}
	\includegraphics[draft=\mydraft, width=0.24\linewidth, trim={600px 100 800px 0}, clip]{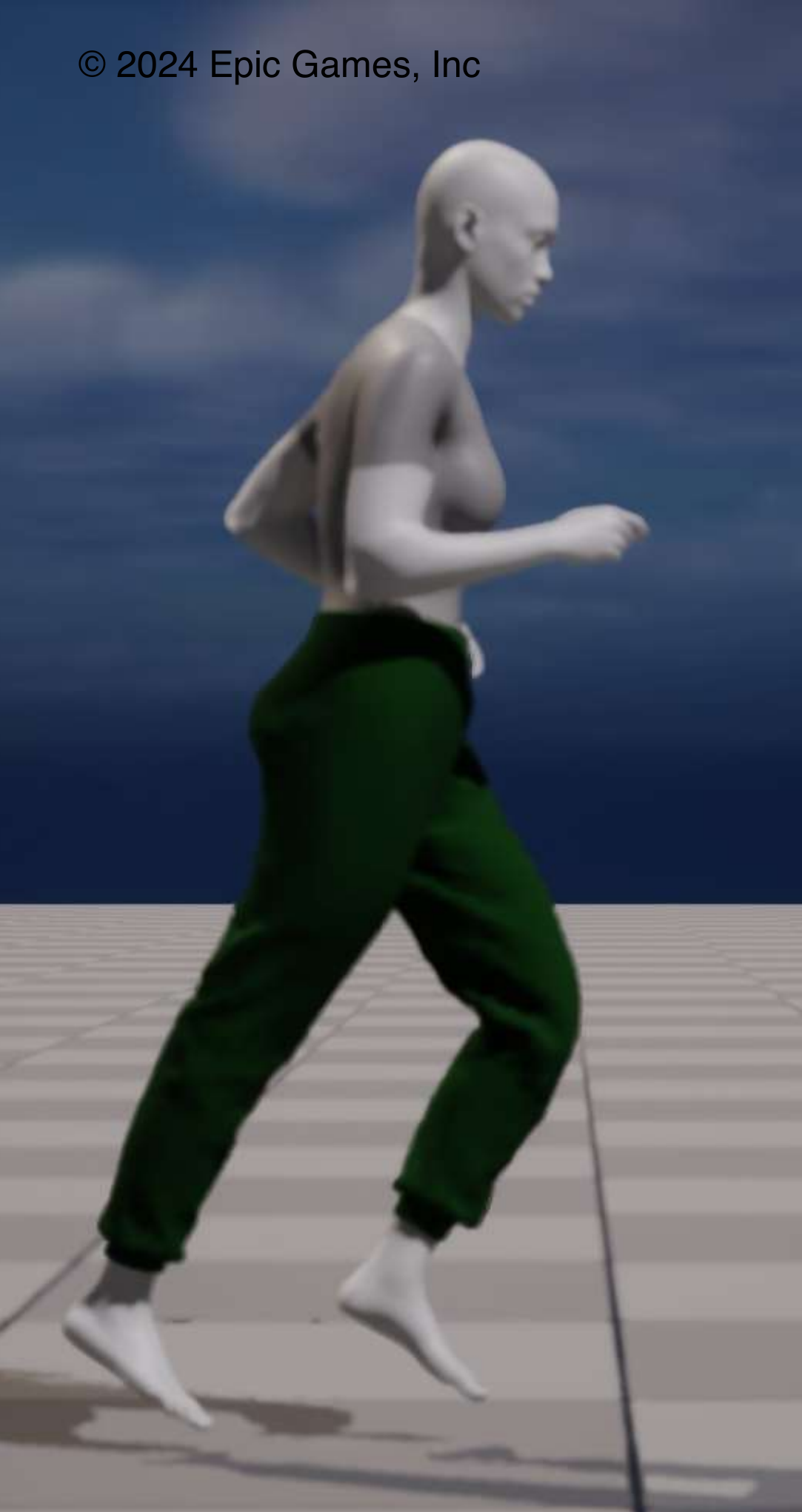}
	\includegraphics[draft=\mydraft, width=0.24\linewidth, trim={600px 100 800px 0}, clip]{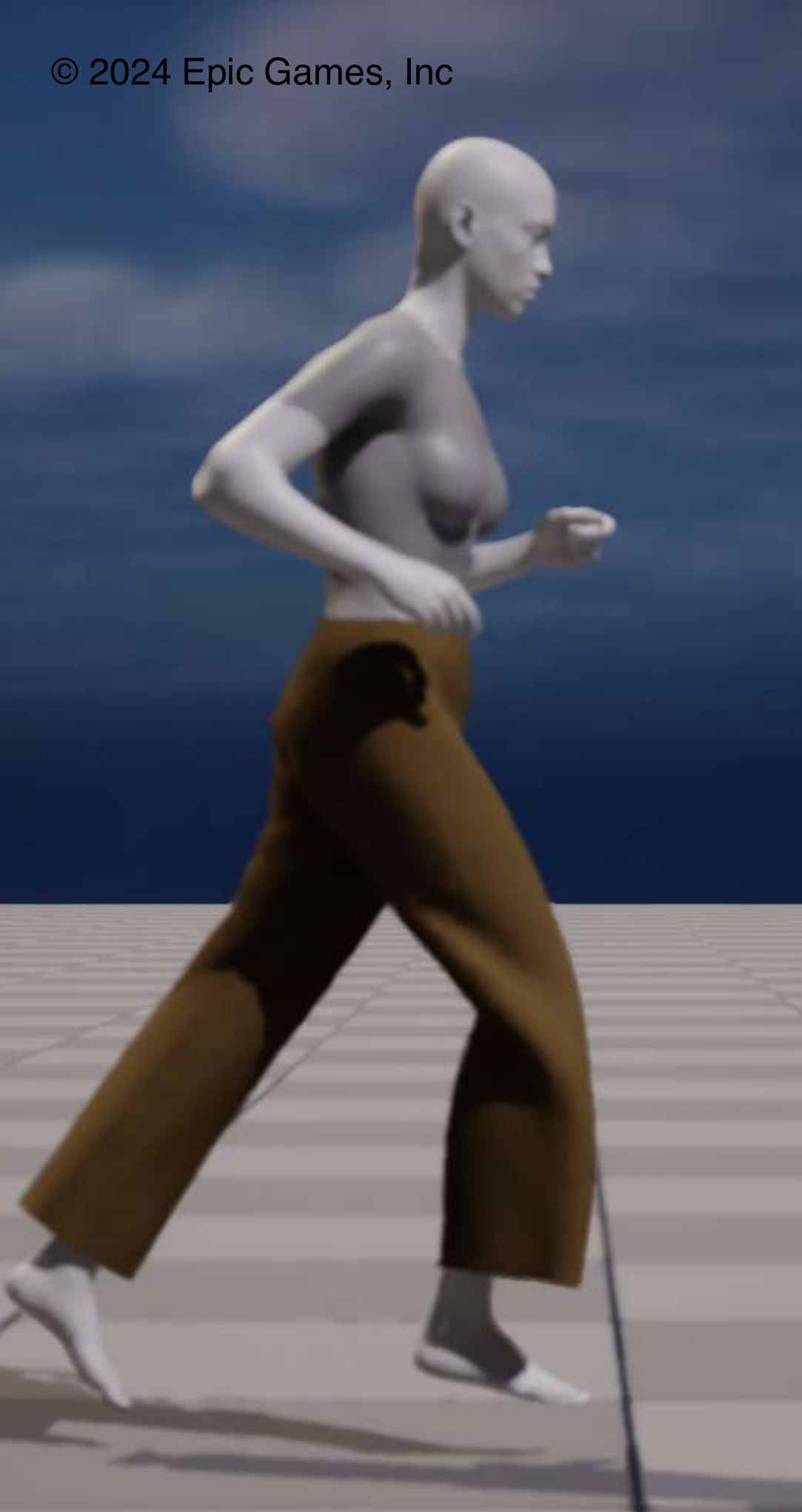}
	\includegraphics[draft=\mydraft, width=0.24\linewidth, trim={600px 100 800px 0}, clip]{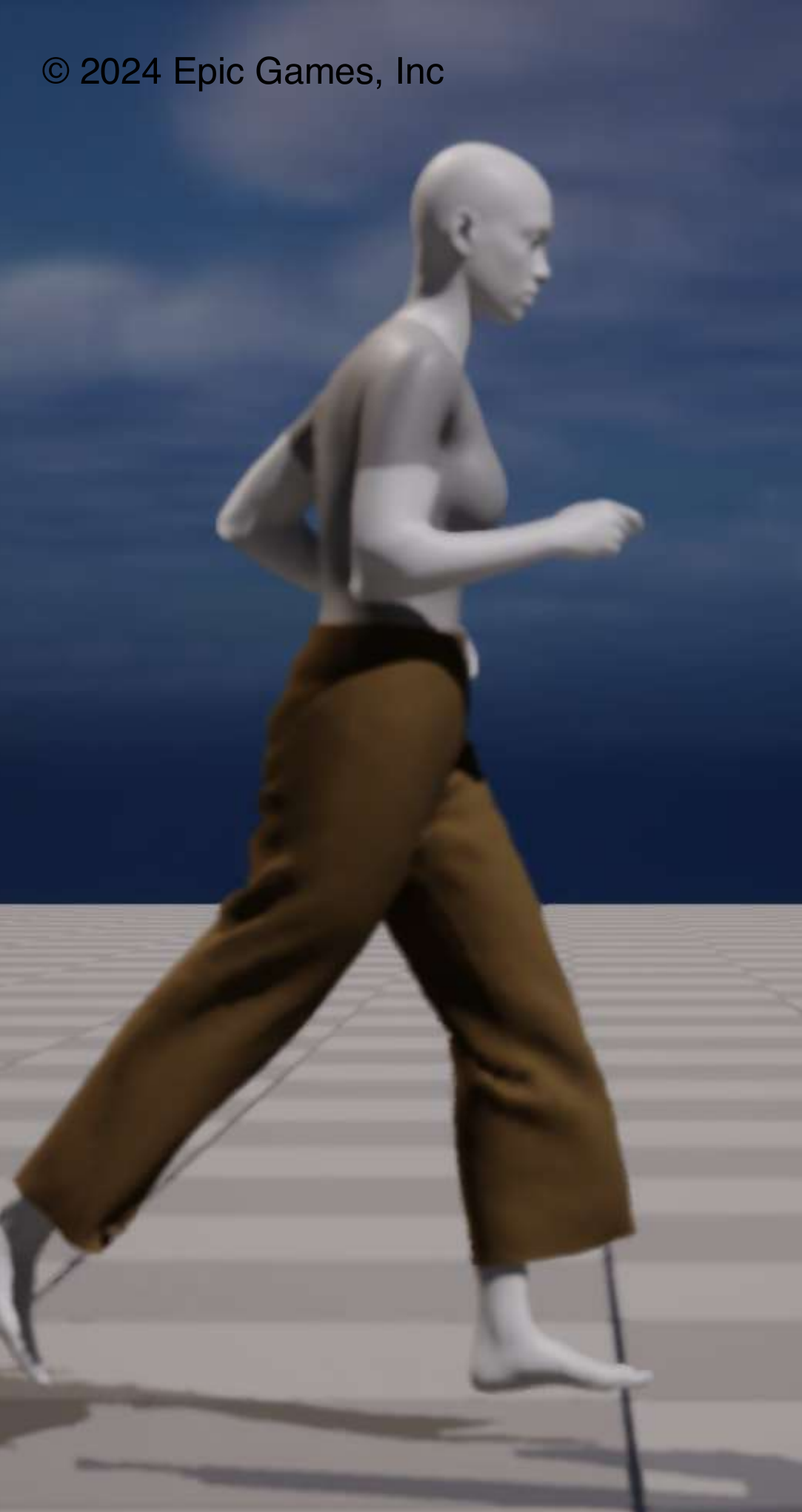}
	
	\includegraphics[draft=\mydraft, width=0.24\linewidth, trim={750px 100 650px 80}, clip]{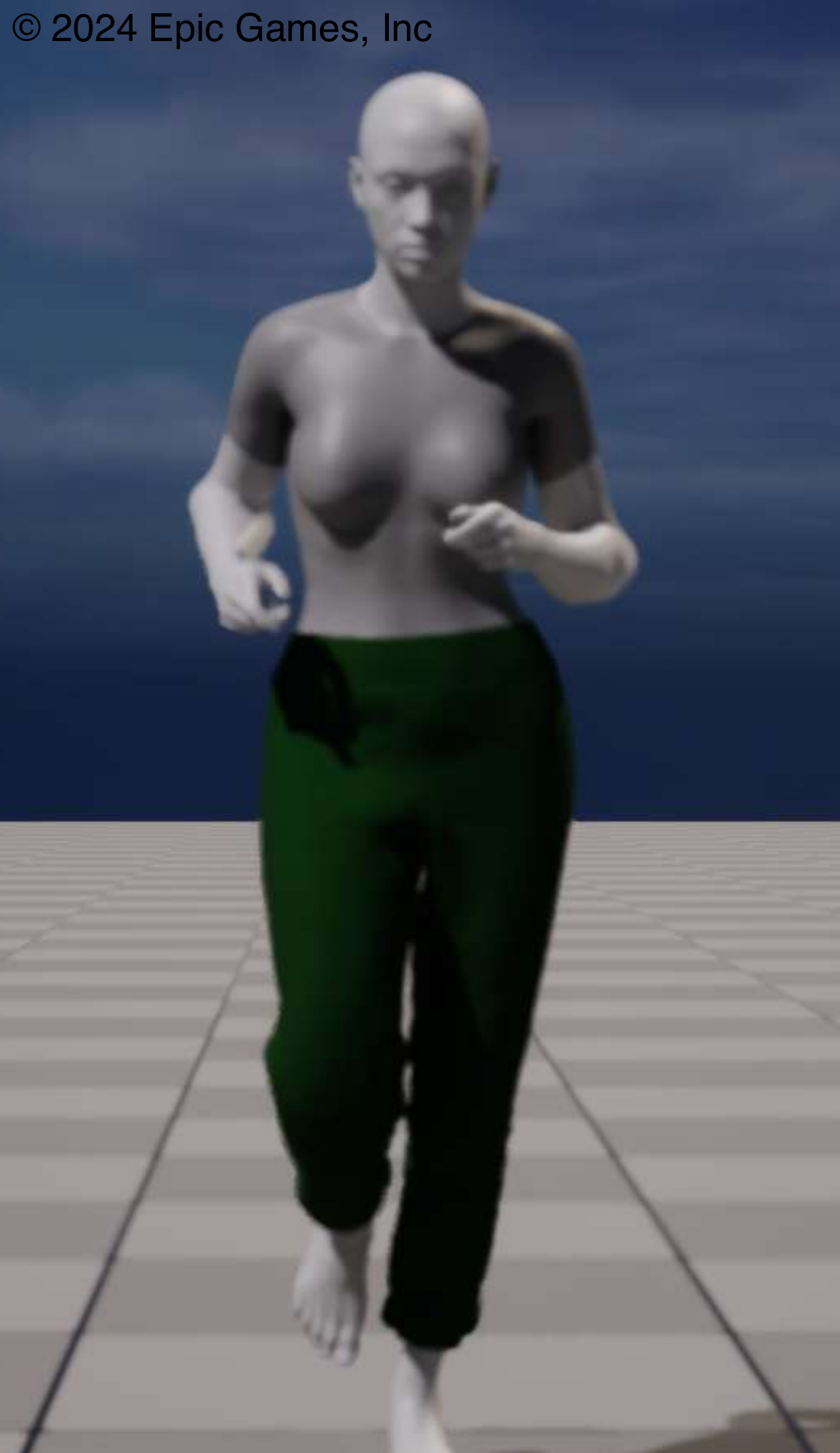}
	\includegraphics[draft=\mydraft, width=0.24\linewidth, trim={750px 100 650px 80}, clip]{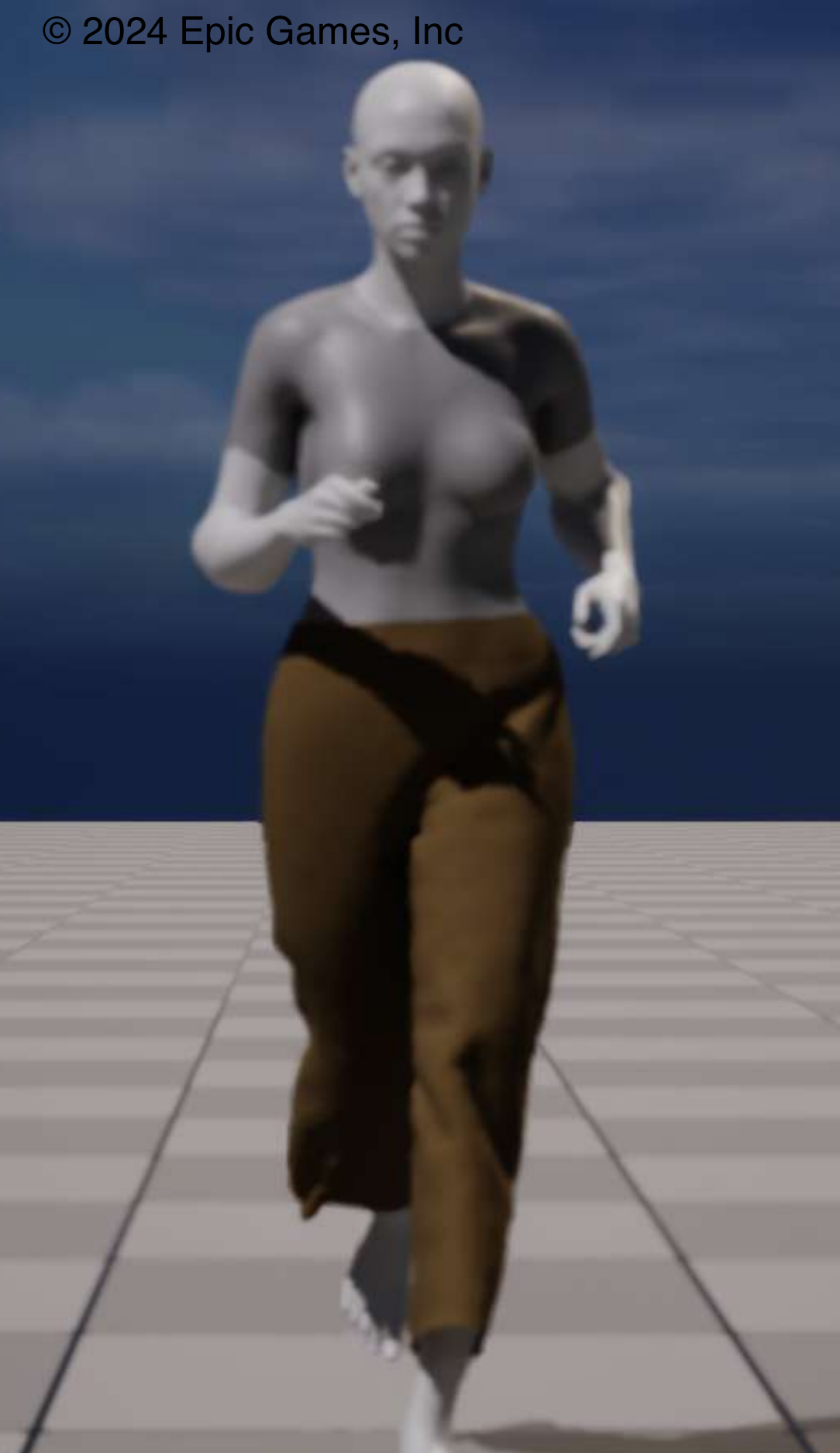}
	\includegraphics[draft=\mydraft, width=0.24\linewidth, trim={550px 90 800px 0}, clip]{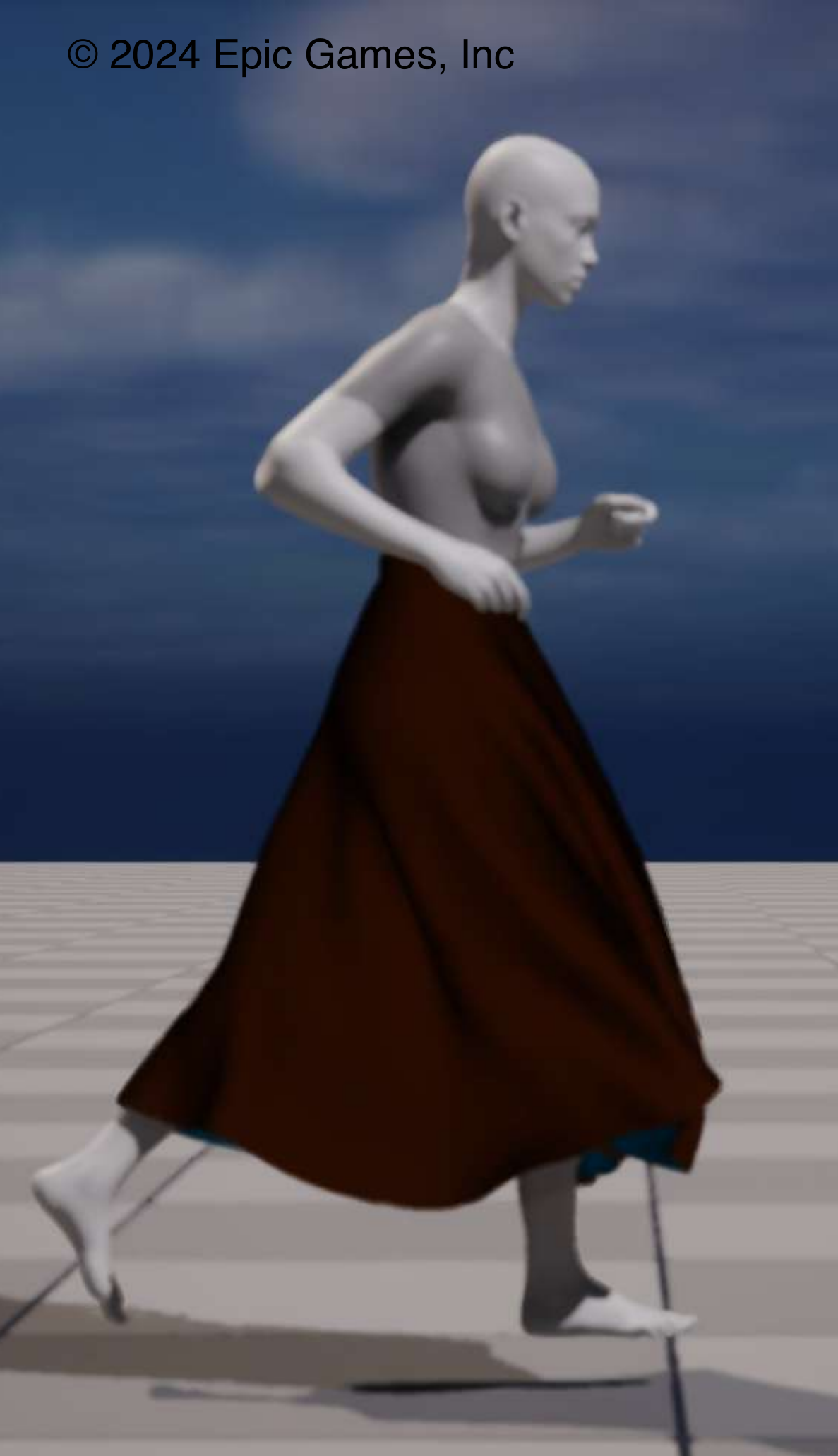}
	\includegraphics[draft=\mydraft, width=0.24\linewidth, trim={550px 90 800px 0}, clip]{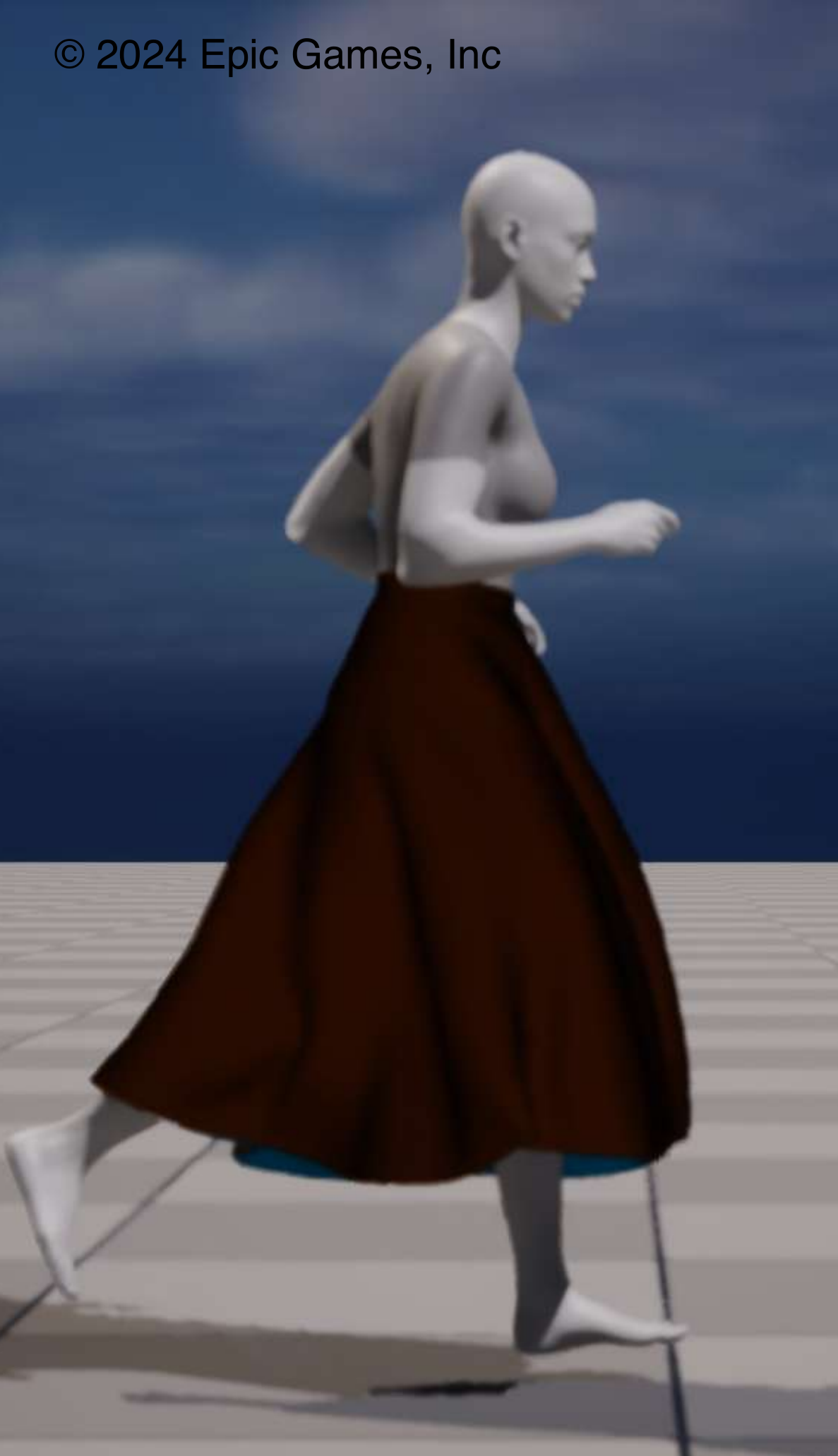}
	\caption{Cloth simulation using our SDFs.}
	\label{fig:cloth_examples}
\end{figure}

We demonstrate the efficacy of our approach in practical cloth simulation examples over different types of garments.
These are show in Figures~\ref{fig:teaser} and \ref{fig:cloth_examples}.
We use the SDF in a standard way to resolve collisions during simulation.
Specifically, the SDF is queried to determine if a cloth particle is inside of the body and push it outwards in negative gradient direction if it is inside.
We use finite forward differencing to compute the gradient normal, which requires 3 more queries (one per direction) per particle inside the body.
For optimization we batch queries whenever possible, because increasing the batch size for the model inference reduces average time cost per particle.
We achieve real-time performance with clothing meshes consisting of $4-6K$ particles.
Collision detection/resolution takes between $10$ to $25$ percent of the total simulation time (See Table~\ref{tbl:inference_times}).\\
\\
We also explicitly illustrate that our learning-based SDFs successfully predict the avatar skin boundary geometry for the any pose in a continuous motion. 
Figure~\ref{fig:iso_vs_gt_W8_d2} shows the 0-levelset of the learned SDFs in a jogging sequence.
The result is remarkably comparable to the ground truth. 
Figure~\ref{fig:lower_body_multiple_epochs} illustrates the effect of training convergence.
Improved geometric detail clearly arises with increased training epochs. 

\begin{table}[t]
	\caption{\textbf{Simulation Timing}}
	\resizebox{\columnwidth}{!}{%
		\begin{tabular}{@{}lllll@{}}
			\toprule
			Example & $N_P$ & $T_{sim}$ & $T_{SDF}$ & $Per_{SDF}$\\ \midrule
			Green Pants &4305& $13.7ms$ & $2.55ms$  &~$19\%$ \\ 
			Yellow Pants &4169& $14.2ms$ & $3.2ms$ &~$22.5\%$\\ 
			Skirt &6056& $37.2$ & 3.78& ~$10\%$\\ 
			\bottomrule
		\end{tabular}
	}
	\label{tbl:inference_times}
	\caption{$N_P$ = Number of particles on the garment cloth. $T_{sim} $ = Simulation time per frame. $T_{SDF}$ = Total time for SDF computation. $Per_{SDF}$ is the percentage of the time used for collision detection using our learned SDF in total simulation.} 
\end{table}

\section{Discussion and Future Work}
Our method allows for real-time SDF queries in practical simulation of clothing collision against deformable avatar skin surfaces.
The collection of shallow networks is essential for achieving efficient evaluation time.
While we show that this can be done accurately in the context of pants and dress collisions with the lower torso, there is still room for improvement.
Increasing the number of weights and biases greatly enhances the expressivity of the network, as shown in Figure~\ref{fig:calf_single_diff_models}.
In future work, we would like to achieve higher accuracy for a given performance constraint.
This can be done with the investigation of novel network architectures.
Lastly, we assume that joint degrees of freedom $\js_i$ will not deform the skin too dramatically as in regions $\Omega_j$ far from the joint.
This will not always be the case and our method could be adjusted to better resolve these cases.

\bibliographystyle{ACM-Reference-Format}
\bibliography{references2}

\clearpage

\end{document}